\PassOptionsToPackage{unicode}{hyperref}
\PassOptionsToPackage{hyphens}{url}
\PassOptionsToPackage{dvipsnames,svgnames,x11names}{xcolor}
\documentclass[
  11pt,
]{article}
\usepackage{xcolor}
\usepackage[margin=1in]{geometry}
\usepackage{amsmath,amssymb}
\usepackage{iftex}
\ifPDFTeX
  \usepackage[T1]{fontenc}
  \usepackage[utf8]{inputenc}
  \usepackage{textcomp} 
\else 
  \usepackage{unicode-math} 
  \defaultfontfeatures{Scale=MatchLowercase}
  \defaultfontfeatures[\rmfamily]{Ligatures=TeX,Scale=1}
\fi
\usepackage{lmodern}
\ifPDFTeX\else
\fi
\IfFileExists{upquote.sty}{\usepackage{upquote}}{}
\IfFileExists{microtype.sty}{
  \usepackage[]{microtype}
  \UseMicrotypeSet[protrusion]{basicmath} 
}{}
\makeatletter
\@ifundefined{KOMAClassName}{
  \IfFileExists{parskip.sty}{%
    \usepackage{parskip}
  }{
    \setlength{\parindent}{0pt}
    \setlength{\parskip}{6pt plus 2pt minus 1pt}}
}{
  \KOMAoptions{parskip=half}}
\makeatother
\usepackage{longtable,booktabs,array}
\usepackage{caption}
\usepackage{calc} 
\usepackage{etoolbox}
\makeatletter
\patchcmd\longtable{\par}{\if@noskipsec\mbox{}\fi\par}{}{}
\makeatother
\usepackage{graphicx}
\makeatletter
\newsavebox\pandoc@box
\newcommand*\pandocbounded[1]{
  \sbox\pandoc@box{#1}%
  \Gscale@div\@tempa{\textheight}{\dimexpr\ht\pandoc@box+\dp\pandoc@box\relax}%
  \Gscale@div\@tempb{\linewidth}{\wd\pandoc@box}%
  \ifdim\@tempb\p@<\@tempa\p@\let\@tempa\@tempb\fi
  \ifdim\@tempa\p@<\p@\scalebox{\@tempa}{\usebox\pandoc@box}%
  \else\usebox{\pandoc@box}%
  \fi%
}
\def\fps@figure{htbp}
\makeatother
\providecommand{\tightlist}{%
  \setlength{\itemsep}{0pt}\setlength{\parskip}{0pt}}
\usepackage[]{natbib}
\usepackage{needspace}
\let\origunderscore\_
\renewcommand{\_}{\origunderscore\allowbreak}
\usepackage{bookmark}
\IfFileExists{xurl.sty}{\usepackage{xurl}}{} 
\makeatletter
\@ifundefined{xmpquote}{}{}
\makeatother
\hypersetup{
  pdftitle={When the next byte's type is fixed: why entropy patching skips computed values under tight budgets},
  pdfauthor={Nicolás Vera Zúñiga --- Independent researcher, Chile --- nicovera@quetru.cl},
  colorlinks=true,
  linkcolor={blue},
  filecolor={Maroon},
  citecolor={blue},
  urlcolor={blue},
  pdfcreator={LaTeX via pandoc}}

\title{When the next byte's type is fixed: why entropy patching skips
computed values under tight budgets}
\author{Nicol\'as Vera Z\'u\~niga\\
Independent Researcher, Chile\\
\texttt{nicovera@quetru.cl}}
\date{}
\date{}

\ifPDFTeX
\DeclareUnicodeCharacter{2265}{\ensuremath{\geq}}
\DeclareUnicodeCharacter{2264}{\ensuremath{\leq}}
\DeclareUnicodeCharacter{00D7}{\ensuremath{\times}}
\DeclareUnicodeCharacter{2192}{\ensuremath{\rightarrow}}
\DeclareUnicodeCharacter{00B5}{\ensuremath{\mu}}
\DeclareUnicodeCharacter{00B0}{\ensuremath{^{\circ}}}
\DeclareUnicodeCharacter{00B1}{\ensuremath{\pm}}
\DeclareUnicodeCharacter{2212}{\ensuremath{-}}
\fi
\begin{document}
\maketitle

\begin{abstract}
The Byte Latent Transformer (BLT) runs a large model once per patch and
starts patches where a small model's next-byte entropy is high. Under
tight budgets this rule skips computed values, such as the number after
\texttt{=} in math. On BLT-1B at inference we trace the miss to the cue
before a value: when the preceding bytes fix the next byte's type, the
patcher's entropy is capped near log2 10 bits, close to a 10\% budget's
threshold. Entropy then starts a patch at 0\% of values after a dollar
sign and 19\% after \texttt{=}; writing \texttt{is} instead of
\texttt{=} raises the 19\% to 66\%. Version 1's two-byte lookup table,
motivated by LZ novelty, acts as a learned detector of cues that tend to
precede new values in its fitting text: on the solutions rewritten as
prose it covers every value after a dollar sign and none after a plain
space, and its margin over entropy falls from +42.1 to -8.9 points,
partly because prose gives entropy more starts per value. In BLT-style
models trained from scratch (1M-50M parameters, 128-byte context, 3
seeds per rule) and scored by teacher-forced exact match on answer
spans, the table beats entropy at 50M by 36.8 points on computed results
and 29.6 on final answers, which mostly copy the last computed result. A
hand-written marker rule is not distinguishable from the table at 1M and
12M; its 50M control was not run. At 1M and 12M the table ran on a T4
GPU in fp16 and the baselines on an A40 in bf16; at 50M all ran on the
A40. End to end, no 1M-50M model solves more than 2 of 100 GSM8K
problems. A label-free probe of BLT-1B's patcher finds prose values,
mostly after a dollar sign, but costs final answers. In six domains, no
rule we tested beat entropy on exact match of whole words and numbers in
more than one domain. The positive findings concern tight budgets on
cue-formatted text. This version corrects version 1; the work is closed,
with no venue intended.
\end{abstract}

\section{1. Introduction}\label{introduction}

Byte-level models avoid a fixed tokenizer by working on raw bytes and
recover efficiency by grouping bytes into variable-length patches
\citep{pagnoni2025blt, slagle2024spacebyte, hwang2025hnet}. In BLT
\citep{pagnoni2025blt}, a small byte-level language model estimates the
entropy of each next byte, a patch starts where that entropy crosses a
threshold, and the large global transformer runs once per patch. The
guiding principle, shared by later work, is that compute should go where
prediction is hard: Scratchpad Patching triggers extra global steps by
next-byte entropy \citep{zheng2026scratchpad}, and SOMBRERO trains
learned chunkers to place boundaries on high-surprisal bytes
\citep{neitemeier2026sombrero}. Fewer patches mean less global compute.
BLT's own scaling runs use average patches of 6 and 8 bytes
\citep[§5.1]{pagnoni2025blt}, so that about 12.5-16.7\% of bytes start a
patch, and at its default threshold BLT-1B starts a patch at 25.5\% of
the bytes of GSM8K solutions \citep{vera2026easy}. We study budgets of
10\% and 15\% of bytes.

Earlier work identified a blind spot in this principle
\citep{vera2026easy}. At a computed result, such as the number after
\texttt{16\ -\ 3\ -\ 4\ =}, the value requires computation over context
the small patcher cannot use, yet the position is not among those the
patcher finds most surprising. Under tight patch budgets entropy spends
patches elsewhere, and accuracy on these positions collapses: in BLT-1B
at a 10\% patch rate, on the 294 GSM8K test problems of §5 and §6,
entropy patching starts a patch at 16\% of computed results and gets
19.0\% of them exactly right. That work proposed a hand-written rule (a
boundary after each \texttt{=}) and \emph{boundary dependence}, a
label-free table of how much the model's loss rises when a patch start
is removed.

We trace the miss to the cue before each value (§5). When the preceding
bytes fix the next byte's type, as \texttt{=} or \texttt{\$} before a
number do, BLT-1B's patcher puts nearly all its probability on the ten
digits, so its entropy stays near or below that of a uniform choice
among them, \(\log_2 10 = 3.32\) bits. That is where the threshold of a
10\% budget sits on our text (3.29-3.36 bits): of the values after
\texttt{=} or \texttt{\$} that entropy misses, 92\% have a digit
probability of at least 0.9 at their first digit, and entropy starts a
patch at none of the values after \texttt{\$}. Two contrasts point to
the cue. Within the same prose texts and at the same budget, entropy
covers 75.5\% of the values that follow a plain space, where the next
byte's type is open, and none of those after \texttt{\$}; and writing
\texttt{is} instead of \texttt{=} before each value raises entropy's
coverage of those values from 18.9\% to 65.7\%. Neither isolates the cue
cleanly: the first compares different values (those after \texttt{\$}
are money), and the swap adds a byte before each value. This is a
ceiling for BLT-1B at tight budgets, not a law for every entropy-gated
method, and it lifts as the budget grows.

Version 1 of this paper took a cheaper route, motivated by compression.
An LZ77 compressor \citep{ziv1977universal} encodes text as copies of
earlier substrings plus new literals, so version 1 started a patch where
the upcoming bytes are unlikely to repeat the recent context, predicted
from the previous two bytes by a table fitted in seconds on solution
text, with no labels and no model. The evidence below shows that the
table works for a simpler reason: it learns cues that tend to precede
new values in its fitting text, here \texttt{=} and \texttt{\$}. On
prose rewrites of the same problems it starts a patch at every value
after \texttt{\$} and at none after a plain space, and falls below
entropy, partly because prose has more bytes per value and so gives
entropy more starts per value. In execution traces, where the same cue
precedes computed and copied values, it starts a patch at neither.
Without a fixed cue it covers 4.9\% of values, on templated text.
Trained from scratch, a hand-written rule with four markers shows no
detectable difference from it at 1M and 12M parameters, with wide
intervals and point estimates that favour the rule. We keep LZ novelty
only as that original motivation.

All accuracies in this paper are teacher-forced exact match on answer
spans. The models we train from scratch are small (1M-50M parameters,
128-byte context) and solve almost no GSM8K problems end to end, and
BLT-1B is used at inference only, with its patch layout changed. The
findings concern tight patch budgets on text whose values follow fixed
cues.

Our contributions:

\begin{enumerate}
\def\labelenumi{\arabic{enumi}.}
\tightlist
\item
  \textbf{Training from scratch} (§4). On one device, the table beats
  entropy at 50M on every seed (computed results +36.8 points, +17.0 to
  +56.5; final answers +29.6, +26.6 to +32.5). At 12M it does so on
  final answers only (+24.8, +14.7 to +35.0), in a comparison across
  GPUs and precisions, and nothing is detectable at 1M. Final answers
  mostly copy the last computed result, and end to end every model is at
  the floor.
\item
  \textbf{The cue, and what the table detects} (§5). The ceiling and the
  \texttt{is} swap that lifts it; prose rewrites of the same problems,
  where the table's margin over entropy falls from +42.1 to -8.9 points;
  templated text without a fixed cue; and cue rules at inference on
  BLT-1B. In training, the table and the hand-written rule are tied at
  1M and 12M (wide intervals), and the 50M control with the table's
  recipe was not run.
\item
  \textbf{Budget and objective, at inference} (§6). On BLT-1B, a model
  trained under entropy patching, entropy needs about 1.75 times the
  table's patch starts for equal computed-result accuracy (interpolated;
  the interpolation alone allows 1.52 to 2.04), and the table needs more
  than 1.45 times entropy's for equal bits per byte. Entropy plus the
  table's top 3\% of positions gains +31.9 points on computed results
  with no detectable whole-text difference in bits per byte, though bits
  per byte move from the solution (-0.021) to the question (+0.032);
  rules were fixed before each run.
\item
  \textbf{Scope and negative results} (§7 and Appendices A-D). In six
  domains, no rule we tested beat entropy on exact match of whole words
  and numbers in more than one domain; a structure-only rule that adds
  starts at word and number starts beat it on bits in five
  (exploratory). A label-free probe on BLT-1B's patcher state finds
  prose values at tight budgets, mostly after \texttt{\$}, a cue the
  rewrite kept, and loses final answers unless fused with entropy (GSM8K
  only, one LLM writer). Patch starts inside verbatim copies break them,
  and a copy veto restores them (BLT-1B with its 512-byte windows
  restored, synthetic IDs). Classic compressors, value uncertainty as a
  training rule, and switching rules after training fail.
\end{enumerate}

Appendices: A the detector, B copies, C outside math, D other negative
results, E a one-seed training pilot, F the work that was not run and
Study A's interim look, G a check of H-Net checkpoints, H where each
rule spends its budget.

This is version 2 of arXiv:2610.22100. It replaces version 1's title and
framing, corrects or withdraws several of its claims, adds the
disclosures above and the results obtained since, and records a line of
work that is now closed, with no venue intended. ``Changes in version
2'' lists the changes and their reasons.

\section{2. Related work}\label{related-work}

\textbf{Patching rules.} BLT \citep{pagnoni2025blt} uses a global
entropy threshold (and an approximate monotonic variant); entropy spikes
as boundaries go back to dynamic pooling \citep{nawrot2023dynamic}.
MEGABYTE \citep{yu2023megabyte} uses fixed-size patches and SpaceByte
\citep{slagle2024spacebyte} patches at word boundaries; H-Net
\citep{hwang2025hnet}, AU-Net \citep{videau2025aunet}, MrT5
\citep{kallini2024mrt5} and ByteSpan \citep{goriely2025bytespan} learn
or derive boundaries in other ways, and Bolmo
\citep{minixhofer2025bolmo} converts trained subword models into
byte-level models. To our knowledge, none of these examines where
boundaries fall on computed values under a tight budget. (Appendix G
reports a check, not reported before, of public H-Net checkpoints on
GSM8K at their native chunk rates, with its caveats.) Scratchpad
Patching \citep{zheng2026scratchpad} adds entropy-triggered global steps
inside fixed patches, Fast BLT \citep{kallini2026fastblt} speeds up
decoding, and EntropyMoE \citep{liu2026entropymoe} routes patches to
experts by entropy. \citet{haltiuk2026disentangling} argues that the
boundary distribution can be changed nearly independently of language
modeling.

\textbf{Lookup patchers.} BLT notes that an entropy model with a small
enough receptive field can be encoded in an efficient lookup table
\citep[§4.2]{pagnoni2025blt}, and its Fig. 3(f) patches by the entropy
of a small byte model with a two-byte context; its code also has an
unused patcher that decides patch starts from a lookup table indexed by
the previous two tokens (\texttt{bytelatent/data/patcher.py}). The
lookup mechanism of our table is therefore not new. BLT's table would
hold entropies; ours holds the LZ novelty target, which on GSM8K acts as
a detector of the format's value cues (§5).

\textbf{Compute per position.} Pause tokens \citep{goyal2024pause} and
filler tokens \citep{pfau2024filler} give a model extra computation
steps that carry no new text; Mixture-of-Depths \citep{raposo2024mod}
routes compute per token with a learned router; Think-at-Hard
\citep{fu2025thinkathard} runs extra latent iterations only on tokens
that a small module predicts to be wrong. These decide how much compute
a token gets; a patching rule decides where the global model takes a
step at all. For static tokenizers, how numbers are segmented affects
arithmetic \citep{singh2024tokenization}.

\textbf{Boundaries and surprisal.} SOMBRERO
\citep{neitemeier2026sombrero} defines \emph{boundary enrichment}, the
mean surprisal at chunk starts relative to all bytes, and trains a
hierarchical byte model to raise it. ByteFlow \citep{deng2026byteflow}
selects boundaries by the coding-rate gain of the encoder's latent
states.

\textbf{Compression and reuse.} Compression and prediction are two sides
of the same coin \citep{deletang2024language}, and compression alone
does not explain what makes a tokenization good
\citep{schmidt2024tokenization}. Description-length criteria segment
text without supervision \citep{kit1999unsupervised}. zip2zip
\citep{geng2025zip2zip} merges recurring tokens into hypertokens with
online LZW, and LLMA \citep{yang2023llma} drafts spans copied from the
context: both make repeats cheap, whereas we ask where the global
model's compute should go.

\textbf{Prior work.} \citet{vera2026easy} documented the blind spot on
BLT-1B and in models trained from scratch, and proposed the hand-written
rule and boundary dependence; we reuse its training pipeline, corpus,
evaluation and BLT-1B setup, and report its rules as baselines. It
already reports the hand-written rule at 53.7M parameters (43.4\% of
computed results, 85.5\% of final answers; two unclipped seeds) and, on
BLT-1B, the \texttt{=} rule and entropy plus dependence (51.8\% and
67.1\% of computed results at a 10\% budget). This paper adds the cue
mechanism, the two-byte table as a cheap learned cue detector, the
budget-by-objective analysis at inference, and the scope and negative
results.

\section{3. Method and setup}\label{method-and-setup}

\textbf{Novelty table.} For a byte stream \(b\), let \(m_t\) be the
length of the longest prefix of \(b_{t:}\) that also starts within the
previous 127 bytes (capped at 16). Novelty is \(-m_t\): high when the
upcoming bytes do not repeat recent context. It depends on future bytes,
so we predict it causally: the table averages \(-m_t\) over each
preceding two-byte context \((b_{t-2}, b_{t-1})\) of a sample of
training text, falling back to one byte, then to the global mean, for
contexts seen fewer than 20 times. A patch starts at \(t\) when the
table value exceeds a threshold set on training text for a target rate.
For the models trained from scratch the table is fitted on 2 MB of GSM8K
and MATH training solutions in their annotated format; for BLT-1B, on
all 7,473 GSM8K training solutions in its stripped format (no MATH);
and, as a variant, on 4 MB sampled from the pretraining corpus (the
\emph{corpus-fitted table}; about 10\% of that corpus is GSM8K and MATH
solutions, see below). A table is fitted on text written as it will be
scored.

\textbf{Baselines.} \emph{Entropy}: BLT's global threshold on next-byte
entropy, from a one-layer, width-32 entropy model trained on the corpus
(models trained from scratch) or from BLT-1B's own 14-layer patcher.
\emph{Dependence}: the boundary-dependence table of
\citet{vera2026easy}. \emph{Hand-written rule} (syntax + entropy): a
patch start after every math-syntax marker (\texttt{=},
\texttt{\textbackslash{}boxed\{}, \texttt{\#\#\#\#},
\texttt{\textgreater{}\textgreater{}}), the rest of the budget by
entropy; this is the rule called the hand-written rule throughout. On
BLT-1B, whose text has no calculator annotations, we use two
hand-written variants instead, the \texttt{=} rule and the syntax rule
(§5).

\textbf{Targets.} A \emph{computed result} (also called a computed
value) is the value right after \texttt{=}: inside the calculator
annotation
(\texttt{\textless{}\textless{}a*b=c\textgreater{}\textgreater{}}) in
text that keeps annotations, and after \texttt{=} in text without them
(its digits only; a \texttt{\$} before it is not part of the value). A
\emph{final answer} is the number after the final-line marker
(\texttt{\#\#\#\#} or ``The final answer is''). In GSM8K the final
answer usually repeats the last number written before it: in 94.6\% of
the 1,319 test solutions (94.2\% of the 660 used to evaluate the models
trained from scratch) the two are equal as strings, with a median of 9
bytes between that number and the \texttt{\#\#\#\#} marker. Under
teacher forcing, final answers therefore test mostly a short copy.

\textbf{Models and data.} BLT-style models (cross-attention pooling, a
global transformer, a one-block local decoder that attends to the
previous 32 bytes, 128-byte context) at 1M (width 128, 4 global layers,
131M training bytes), 12M (320, 8 layers, 0.5B bytes) and 50M (576, 12
layers, 1.5B bytes), all trained from scratch on the same 1.9 GB corpus
(OpenWebMath \citep{paster2024openwebmath} shards plus GSM8K
\citep{cobbe2021gsm8k} and MATH \citep{hendrycks2021math} training
solutions repeated 20 times to make 10\% of bytes; SHA-256 pinned;
documents overlapping the evaluation set removed). The models therefore
see each training solution about 1.4, 5 and 16 times at 1M, 12M and 50M;
evaluation uses held-out test problems. For comparison, BLT-1B has a
9-layer local decoder and a 14-layer patcher (its Hugging Face
configuration). Each rule's threshold targets 10\% of training bytes
(13\% for the corpus-fitted table), which gives 10.9-12.1\% of
evaluation bytes (entropy 11.2\%, dependence 10.9\%, hand-written rule
12.1\%, table 11.9\%, corpus-fitted table 11.4\%).

\textbf{Hardware and recipe.} At 1M and 12M the table runs (and the
value-uncertainty runs of Appendix D) trained on Kaggle T4 GPUs in fp16
with loss scaling, and the entropy, dependence and hand-written runs on
an NVIDIA A40 in bf16. At 50M every run used the A40. Every 1M and 12M
comparison between the table and a baseline therefore crosses devices
and precisions. The size of that effect was not measured; as a scale for
run-to-run variation, re-training two 12M table seeds, again on T4s,
moved their final answers by -3.3 and +7.1 points. The 1M and 12M runs
are unclipped; the 50M runs in the main comparison use gradient clipping
at norm 1.0. For entropy, clipping had no detectable effect on answers
at 3 seeds (final answers +0.9 points, -2.4 to +4.2; computed results
+6.8, -14.2 to +27.8) but lowered bits per byte (1.439 to 1.416; -0.024,
-0.042 to -0.005). For dependence there was no detectable difference on
any of the three (bits per byte +0.000, -0.017 to +0.018).

\textbf{Evaluation of the models trained from scratch.} Teacher-forced
exact match: greedy decoding of each target given the true preceding 127
bytes, with patch boundaries decided online by the rule from the bytes
seen so far, on held-out GSM8K and MATH solutions (annotations kept),
for 1,000 computed results and 660 final answers (the first 660 GSM8K
test problems), plus bits per byte. End to end (greedy generation from
the question), no run in Table 1 solves more than 1 of 100 GSM8K test
problems (no 1M-50M run in this paper more than 2; Study A's 145M models
have no reported end-to-end score), and MATH
\texttt{\textbackslash{}boxed\{\}} answers are 1.9-4.0\% in every Table
1 run except the diverged 1M seed (0.0\%). We score answer spans with
teacher forcing because end-to-end accuracy is at the floor. Three seeds
per configuration; differences with Welch 95\% intervals over seeds (t
quantiles at the floored Welch-Satterthwaite degrees of freedom; seed
variance only). One 1M dependence run never learned (bits per byte
4.718, against 2.08-2.26 for the other 1M runs) and is excluded by the
exclusion rule of \texttt{scale/analyze.py} (bits per byte above the
median of its size plus 1).

\textbf{BLT-1B.} The Hugging Face conversion of facebook/blt-1b
(\texttt{itazap/blt-1b-hf}), with the 512-byte sliding window restored
by an upstream fix that is not yet merged \citep{vera2026easy}; this is
not the official BLT code. Its default layout starts a patch at 25.5\%
of GSM8K bytes, so 10-15\% budgets lie outside what it was trained with.
End to end it solves about 2\% of GSM8K problems even after adaptation
(released conversion) \citep{vera2026easy}, so every BLT-1B accuracy
here is teacher-forced. Unless stated otherwise: 294 GSM8K test problems
written without annotations and ending ``The final answer is N'' (796
computed results); layouts changed at inference, each rule's threshold
fitted on 300 training problems to a 10\% or 15\% patch rate;
differences with paired bootstrap intervals over problems (2,000
resamples unless stated). Other tests use other problem sets, named
where they appear: 1,292 test solutions for the patcher's entropy (§5)
and the selection analysis (Appendix H); the first prose test, 221 test
problems with 711 computed values (§5, Appendix A); 915 later test
problems with 2,448 computed values (§5, Appendix A); and 60 documents
per domain (Appendix C). Entropy's coverage of computed values at a 10\%
budget therefore appears as 16\% (the 294 problems), 18.4\% (the 1,292),
13.6\% (the 221) and 18.9\% of the values after \texttt{=} (the 915).
Analyses not planned before their data are marked as such.

\needspace{30\baselineskip}

\section{4. Results in math: models trained from
scratch}\label{results-in-math-models-trained-from-scratch}

\textbf{Table 1.} Teacher-forced exact match (\%) on computed results
and final answers, and bits per byte, of BLT-style models trained from
scratch at a \textasciitilde11\% evaluation patch rate. Mean ± sd over 3
seeds. 1M and 12M unclipped, 50M clipped. The device row gives the GPU
and precision of each column's runs: at 1M and 12M the table columns ran
on a T4 in fp16 and the others on an A40 in bf16; at 50M every run used
the A40.

{\def\LTcaptype{none} 
{\footnotesize\setlength{\tabcolsep}{3pt}
\begin{longtable}[]{@{}
  >{\raggedright\arraybackslash}p{(\linewidth - 12\tabcolsep) * \real{0.1047}}
  >{\raggedright\arraybackslash}p{(\linewidth - 12\tabcolsep) * \real{0.1047}}
  >{\raggedright\arraybackslash}p{(\linewidth - 12\tabcolsep) * \real{0.1512}}
  >{\raggedright\arraybackslash}p{(\linewidth - 12\tabcolsep) * \real{0.1860}}
  >{\raggedright\arraybackslash}p{(\linewidth - 12\tabcolsep) * \real{0.1512}}
  >{\raggedright\arraybackslash}p{(\linewidth - 12\tabcolsep) * \real{0.1512}}
  >{\raggedright\arraybackslash}p{(\linewidth - 12\tabcolsep) * \real{0.1512}}@{}}
\toprule\noalign{}
\begin{minipage}[b]{\linewidth}\raggedright
size
\end{minipage} & \begin{minipage}[b]{\linewidth}\raggedright
\end{minipage} & \begin{minipage}[b]{\linewidth}\raggedright
entropy
\end{minipage} & \begin{minipage}[b]{\linewidth}\raggedright
dependence
\end{minipage} & \begin{minipage}[b]{\linewidth}\raggedright
hand-written rule
\end{minipage} & \begin{minipage}[b]{\linewidth}\raggedright
\textbf{novelty table}
\end{minipage} & \begin{minipage}[b]{\linewidth}\raggedright
corpus-fitted table
\end{minipage} \\
\midrule\noalign{}
\endhead
\bottomrule\noalign{}
\endlastfoot
all sizes & eval rate & 11.2\% & 10.9\% & 12.1\% & 11.9\% & 11.4\% \\
1M & device & A40 bf16 & A40 bf16 & A40 bf16 & T4 fp16 & -- \\
& computed & 4.4 ± 0.7 & 4.2 ± 0.1\textsuperscript{a} & 4.6 ± 1.1 & 3.7
± 0.6 & -- \\
& final & 7.0 ± 2.0 & 5.4 ± 1.2\textsuperscript{a} & 45.5 ± 17.5 & 12.6
± 10.5 & -- \\
& bits/byte & 2.093 ± 0.012 & 2.196 ± 0.096\textsuperscript{a} & 2.109 ±
0.014 & 2.125 ± 0.018 & -- \\
12M & device & A40 bf16 & A40 bf16 & A40 bf16 & T4 fp16 & T4
fp16\textsuperscript{c} \\
& computed & 5.8 ± 0.9 & 7.7 ± 2.4 & 12.9 ± 2.4 & 10.1 ± 3.9 & 10.2 ±
2.7 \\
& final & 46.6 ± 4.0 & 56.7 ± 2.4 & 76.3 ± 2.3 & 71.5 ± 0.7 & 69.8 ±
1.4 \\
& bits/byte & 1.657 ± 0.013 & 1.673 ± 0.031 & 1.645 ± 0.022 & 1.674 ±
0.029 & 1.619 ± 0.003 \\
50M & device & A40 bf16 & A40 bf16 & not run\textsuperscript{b} & A40
bf16 & A40 bf16 \\
& computed & 20.7 ± 7.9 & 35.1 ± 1.8 & not run\textsuperscript{b} & 57.5
± 1.1 & 57.3 ± 2.3 \\
& final & 59.9 ± 1.3 & 78.2 ± 0.9 & not run\textsuperscript{b} & 89.5 ±
1.0 & 86.4 ± 1.1 \\
& bits/byte & 1.416 ± 0.009 & 1.420 ± 0.007 & not run\textsuperscript{b}
& 1.446 ± 0.005 & 1.436 ± 0.009 \\
\end{longtable}
}
}

\textsuperscript{a} Two seeds: the third never learned and is excluded,
as in every interval (version 1 averaged it in, giving 2.8, 3.6 and
3.037). \textsuperscript{b} Not run with clipping; two unclipped seeds
are reported in \citet{vera2026easy} (43.4 ± 9.3 computed, 85.5 ± 0.3
final, 1.431 ± 0.014 bits per byte). \textsuperscript{c} At 12M the
corpus-fitted table was also trained at 10\% of training bytes; it then
starts patches at only 8.7\% of evaluation bytes (final answers
66.4-68.0\%, computed results 6.9-9.4\% over 3 seeds), so the 13\%
version is the one matched on evaluation rate.

\textbf{The table against entropy.} At 1M no rule differs reliably from
entropy (the table's final-answer difference is +5.6 points, interval
-20.8 to +32.0; the hand-written rule's +38.4, -5.2 to +82.1). At 12M
the table's final answers are 24.8 points above entropy's (+14.7 to
+35.0), but the two arms ran on different GPUs and precisions (§3), and
its computed-result difference is not distinguishable (+4.3, -5.6 to
+14.1). At 50M, on one device and recipe, the table beats entropy on
every seed: +36.8 points on computed results (+17.0 to +56.5) and +29.6
on final answers (+26.6 to +32.5); it also beats boundary dependence, by
+22.4 (+18.5 to +26.3) and +11.3 (+8.9 to +13.7). Entropy's computed
accuracy varies widely across seeds (12.1, 27.5 and 22.6\%), but its
best seed is far below the table's worst (56.3\%). The final-answer gap
moves from +24.8 at 12M (unclipped, across devices) to +29.6 at 50M
(clipped, one device). With recipe and device changing between sizes,
these results do not show an effect that increases with model size
(Figure 1).

\begin{figure}
\centering
\includegraphics[width=0.95\linewidth,height=\textheight,keepaspectratio,alt={Teacher-forced
exact match of the models trained from scratch, by size, at a
\textasciitilde11\% evaluation patch rate. Lines join the means over 3
seeds; markers are seeds, hollow for runs on a T4 in fp16 and filled for
runs on the A40 in bf16. 1M and 12M unclipped, 50M clipped: recipe and
device change between sizes, so this is not a clean scaling curve. The
hand-written rule was not trained at 50M with clipping; the failed 1M
dependence seed is left out. Drawn by paper2/fig1\_v2.py from
results/scale/runs/.}]{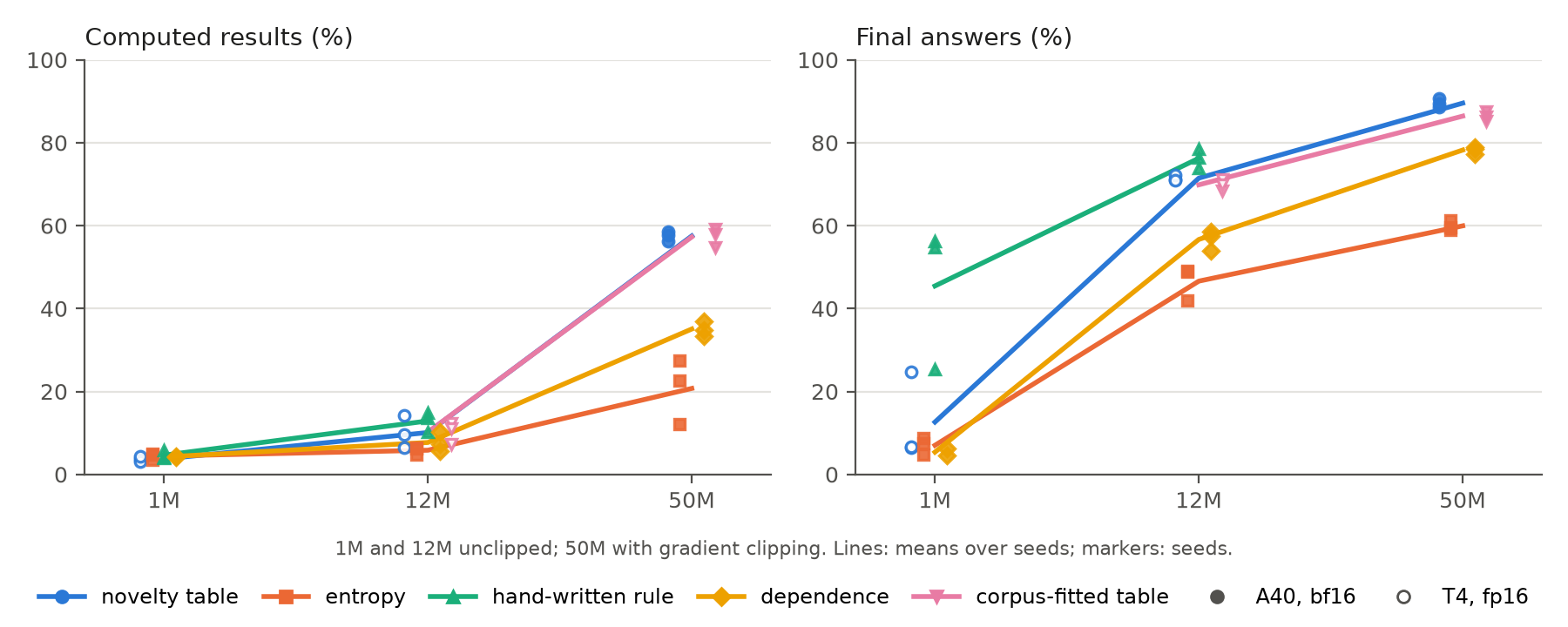}
\caption{Teacher-forced exact match of the models trained from scratch,
by size, at a \textasciitilde11\% evaluation patch rate. Lines join the
means over 3 seeds; markers are seeds, hollow for runs on a T4 in fp16
and filled for runs on the A40 in bf16. 1M and 12M unclipped, 50M
clipped: recipe and device change between sizes, so this is not a clean
scaling curve. The hand-written rule was not trained at 50M with
clipping; the failed 1M dependence seed is left out. Drawn by
\protect\texttt{paper2/fig1\_v2.py} from
\protect\texttt{results/scale/runs/}.}
\end{figure}

\textbf{Final answers.} In the evaluation text of these models, entropy
already starts a patch at every final-answer start (where the small
patcher's surprisal averages 5.39 bits) and dependence at none. Yet at
50M final-answer accuracy follows coverage of computed-result starts,
measured at the same thresholds (Table H1, Appendix H): entropy covers
0.7\% of them and gets 59.9\% of final answers, dependence 36.7\% and
78.2\%, the corpus-fitted table 99.3\% and 86.4\%, the table 100\% and
89.5\%. Since the final answer usually repeats the last number a few
bytes earlier (§3), the final-answer half of these results reads mostly
as getting that number right and copying it, not as a boundary at the
answer.

\textbf{The hand-written rule.} On one device (the A40), the
hand-written rule beats entropy at 12M by +29.6 points on final answers
(+21.2 to +38.1) and +7.1 on computed results (+0.8 to +13.3). Against
the table it shows no detectable difference, with the rule ahead on
point estimates and wide intervals: the table minus the rule is -4.8
points on final answers at 12M (-10.7 to +1.1) and -2.8 on computed
results (-11.1 to +5.5), and -32.8 on final answers at 1M (-70.2 to
+4.5). Both comparisons cross devices. At 50M the rule was never trained
with the table's recipe (clipped, 3 seeds): that test was filed before
any run, and an amendment before any data made final answers decisive,
but it was not run (Appendix F). Two earlier unclipped runs are public
in \citet{vera2026easy} (43.4\% computed, 85.5\% final). The clipped
table minus these runs is +4.0 points on final answers (+1.4 to +6.7)
and +14.2 on computed results (-69.5 to +97.8): a preview with a
different recipe and seed count, not a test. Against unclipped entropy,
on the same recipe, the rule gains +26.5 points on final answers (+23.1
to +29.9).

\textbf{The corpus-fitted table keeps most of the gain.} Fitted on a
sample of the pretraining corpus instead of the solutions alone, the
table shows no detectable difference from the solution-fitted one on
computed results at 50M (57.3\% vs 57.5\%; -0.2 points, -6.6 to +6.1)
and is 3.1 points lower on final answers (-5.9 to -0.4); its bits per
byte show no detectable difference either (-0.011, -0.030 to +0.008).
Because about 10\% of the corpus is GSM8K and MATH solutions, this does
not show independence from in-domain text; a table fitted on OpenWebMath
alone was not tested. At inference on BLT-1B, whose text has no
calculator annotations, the corpus-fitted table starts a patch at none
of the computed results and is 8.8 points below entropy (-11.4 to -6.2;
Table 2): a table must be fitted on text written as it will be scored.

\textbf{Cost in bits per byte.} At 50M the table's bits per byte are
higher than entropy's (1.446 vs 1.416; +0.031, +0.012 to +0.049),
although it starts patches at more evaluation bytes (11.9\% vs 11.2\%);
at a matched rate the gap would likely be larger. The equal-rate
comparison was planned and not run (Appendix F).

\textbf{Study A (145M) was stopped.} A preregistered comparison of the
table with entropy at 145M parameters (called ``150M'' in its
preregistration) was stopped when this line of work closed, before two
of its three seeds per rule had finished, so none of its predictions was
evaluated. Its one interim look, at seed 0, is a single seed on other
hardware than 50M (Kaggle T4s in fp16), interim and not confirmatory,
and both runs' bits per byte were worse than every 50M run, which fired
the look's rule for broken training. Its accuracies therefore cannot be
read as a scale result. Appendix F reports the look with its caveats and
the amendments that set the design.

\section{5. Why entropy skips cued values, and what the table
detects}\label{why-entropy-skips-cued-values-and-what-the-table-detects}

In worked math, values follow fixed cues such as \texttt{=} and
\texttt{\$}. This section shows why entropy misses values after such
cues under a tight budget, and that the table detects the cues rather
than novelty. Everything here is measured at inference on BLT-1B, at a
10\% budget unless stated.

\textbf{Where the threshold falls.} On 1,292 GSM8K test solutions,
computed-result starts have above-average entropy under BLT-1B's patcher
(mean 2.90 bits, against 1.11 over all bytes; mean percentile 85.5, 85.2
to 85.7), but only 18.4\% of them (17.4 to 19.5) pass the
90th-percentile threshold that a 10\% budget sets (3.31 bits). They
receive 1.3\% of that budget; word starts receive 20.8\%, and positions
that start no labelled unit, including the whole question, 69.2\%. The
reason proposed here is the cue. When the preceding bytes fix the next
byte's type, as \texttt{=} or \texttt{\$} before a number do, the
patcher puts nearly all its probability on the ten digits, so its
entropy cannot rise much above that of a uniform choice among them,
\(\log_2 10 = 3.32\) bits, which is about where the threshold lies.

\textbf{The same problems written as prose} (rules fixed before the
run). We had a language model (Claude Opus 5.5, one fixed prompt; data
card in ``Code and data'') rewrite GSM8K solutions as natural
explanatory prose: no \texttt{=} and no arithmetic symbol between
numbers, every computed result of the original kept in order with the
same digits, a \texttt{\$} before money allowed, and the same final
line. On the 221 test problems with computed values (711 values), with
thresholds and tables refitted on each writing's own training problems,
entropy's coverage of the values rises from 13.6\% to 57.2\% (+43.6
points, +38.8 to +48.2). A start at each value, against none, still pays
in prose (exact match from 6.2\% to 34.6\%; +28.4, +24.7 to +32.2), but
less than in the original (from 9.3\% to 53.6\%, +44.3; difference
-15.9, -20.8 to -11.3). The table's margin over entropy falls from +42.1
points (+37.7 to +46.4) in the original to -8.9 (-14.0 to -3.8) in
prose, where it covers 22.9\% of the values. By the decision rule fixed
before the run the result is ``format-tied'' (the margin depends on the
writing). The values are also harder in prose: BLT-1B's own layout gets
77.9\% of them in the original and 56.8\% in prose. Split by the bytes
before each value (after the data), entropy covers 17.8\% of the
original's values after \texttt{=} and none after \texttt{\$}, and
74.3\% of the prose values after a plain space and again none after
\texttt{\$}. The table starts a patch at every value after \texttt{\$}
in both writings (167 and 163 values) and at every value after
\texttt{=}, but at none of the 548 prose values after a plain space.

\textbf{A test of the ceiling} (rules filed before its data). The split
was then tested on the 915 later GSM8K test problems of Appendix A
(2,448 computed values). Their coverage by cue and the patcher's digit
probabilities had not been computed in any writing, and the \texttt{is}
writing was new; their prose rewrites had been scored for exact match in
Appendix A's tests, and in the original writing they are among the 1,292
solutions above (pooled), so their exact-match results by cue had been
seen. Each problem was scored in three writings: the original format,
its prose rewrite, and the original with \texttt{is} written instead of
\texttt{=} before each value. Thresholds were fitted on training
problems: 3.36 bits for a 10\% budget in the original format and 3.29 in
prose. Of the values after \texttt{=} or \texttt{\$} that entropy misses
(2,672 missed values in both writings), 92.2\% (90.8 to 93.5) have a
digit probability of at least 0.9 at their first digit. After
\texttt{\$}, the patcher's entropy at the first digit has a median of
2.76 bits and a 99th percentile of 3.13, and entropy starts a patch at
none of the values, in either writing (586 and 575 values). After
\texttt{=} it starts a patch at 18.9\% of 1,862 values; in prose, after
a plain space, where the type is open, at 75.5\% of 1,873 (+75.5 points
over the \texttt{\$} values of the same texts, +73.5 to +77.6). Writing
\texttt{is} instead of \texttt{=} raises entropy's coverage of those
values from 18.9\% to 65.7\% (+46.9, +44.5 to +49.3), and after the swap
coverage follows the digit probability: 11.9\% where it is still at
least 0.9, 81.5\% where it is lower. A start at the value adds 34.3
points of exact match on prose values after \texttt{\$}, against 4.5
after a plain space (difference +29.7, +25.4 to +34.3); this split had
been seen before the rule was written, so it is reported, not tested.

\textbf{The ceiling is a tight-budget effect.} As the budget grows, the
threshold falls below \(\log_2 10\) and the values after \texttt{\$}
become visible: entropy covers 0\%, 33.1\%, 85.3\% and 96.9\% of them at
budgets of 10\%, 15\%, 20\% and 25\% of bytes (original format), and
BLT-1B's own layout starts a patch at 25.5\% of bytes. Budget per value
matters too: prose has more bytes per value, so the same 10\% budget
gives it more starts per value. The filed decision rule reads this as
``mixed'', neither a pure cue effect nor a pure budget effect. At 10\%,
entropy covers 14.3\% of the original's values and 57.8\% of the prose
values, a 43.4-point gap. At prose's budget per value (an 18.5\% rate,
threshold 2.52 bits), the original format covers 82.8\% of its values,
more than prose covers at 10\%; cut to the original's budget per value
(a 5.4\% rate), prose covers 30.2\%, which closes 63\% of the 43.4-point
gap. The within-text contrast above (0\% after \texttt{\$} against
75.5\% after a plain space, at one budget) is the part of the cue effect
that budget per value cannot explain. We state the ceiling for BLT-1B at
tight budgets, not as a law for every entropy-gated method.

\textbf{Templated text without a fixed cue.} We also generated
GSM8K-style problems whose computed values are introduced by varied
phrasing (``Adding 11 to 17 gives 28'', ``Iris ends up with 47 boxes'')
and never by \texttt{=}, and refitted the table on 3,000 of them. On 300
test problems (1,156 values; value starts are 1.1\% of bytes), the table
covers 4.9\% of the values, about half of what a random 10\% layout
would. The true LZ novelty target \(m_t\), computed with look-ahead,
covers 33.8\%, and entropy 61.2\%: in this templated text the words are
formulaic and predictable, so entropy's budget falls on the numbers.
Without a fixed cue, the two-byte table finds fewer of the values than
entropy and far fewer than its own target. Accuracy cannot be compared
here: BLT-1B gets 3.5\% of these values right under entropy and 4.2\%
with a start forced at every value.

\textbf{Cue rules at inference.} On BLT-1B, layouts can be changed at
inference without retraining (Table 2). In this text the \texttt{=} rule
forces a start at the first digit of every labelled computed result
(after \texttt{=} or \texttt{=\ \$}) and fills the rest of the budget by
entropy. The labels are defined by the same \texttt{=}, so its 100\%
coverage holds by construction and it acts as a label-position oracle;
the syntax rule adds a start after ``The final answer is''. At 10\% the
table also covers every computed-result start and gets 56.5\% of them
right, against 51.8\% for the \texttt{=} rule (+4.8 points, +0.9 to
+8.5) and 19.0\% for entropy; since both cover every result, the
difference comes from where the table places its other boundaries. At
15\% the table and the \texttt{=} rule show no detectable difference
(+0.4, -3.2 to +4.1). A SpaceByte-style rule, entropy restricted to
positions after a non-alphanumeric byte, helps little at 10\% (22.2\%)
and much at 15\% (61.1\%; the table is ahead by +5.4, +1.4 to +9.0).
Entropy plus boundary dependence \citep{vera2026easy}, another
label-free two-byte table, scores higher than the table (67.1\% at 10\%
and 76.5\% at 15\%, against 56.5\% and 66.5\%; no paired interval
computed here). Both are lookups at inference; the novelty table is only
cheaper to fit (solution text in the scored format; no model, losses or
labels). Alone, the table never starts a patch at the final answer,
which here follows ``The final answer is'' rather than
\texttt{\#\#\#\#}, and loses final answers (19.4\% against 51.4\% for
entropy at 10\%); it also loses whole-text bits per byte (1.042 against
0.867; +0.175, +0.164 to +0.186). With a forced start at the answer
marker (10.4\% of bytes) it gets 54.8\% of computed results and 65.6\%
of final answers. The corpus-fitted table, fitted on annotated text,
covers none of the computed results here.

\needspace{26\baselineskip}

\textbf{Table 2.} BLT-1B at inference, a model trained under entropy
patching, with only the layout changed (Hugging Face conversion,
512-byte window restored); 294 GSM8K test problems without annotations,
796 computed results, teacher-forced. Patch rate on the test text,
coverage of computed-result starts, and exact match (\%) on computed
results and final answers. Dependence rows from \citet{vera2026easy}.
Value uncertainty and classic compressors as patch rules are in Appendix
D.

{\def\LTcaptype{none} 
{\footnotesize\setlength{\tabcolsep}{3pt}
\begin{longtable}[]{@{}
  >{\raggedright\arraybackslash}p{(\linewidth - 8\tabcolsep) * \real{0.6154}}
  >{\raggedright\arraybackslash}p{(\linewidth - 8\tabcolsep) * \real{0.0769}}
  >{\raggedright\arraybackslash}p{(\linewidth - 8\tabcolsep) * \real{0.1077}}
  >{\raggedright\arraybackslash}p{(\linewidth - 8\tabcolsep) * \real{0.1231}}
  >{\raggedright\arraybackslash}p{(\linewidth - 8\tabcolsep) * \real{0.0769}}@{}}
\toprule\noalign{}
\begin{minipage}[b]{\linewidth}\raggedright
layout
\end{minipage} & \begin{minipage}[b]{\linewidth}\raggedright
rate
\end{minipage} & \begin{minipage}[b]{\linewidth}\raggedright
covered
\end{minipage} & \begin{minipage}[b]{\linewidth}\raggedright
computed
\end{minipage} & \begin{minipage}[b]{\linewidth}\raggedright
final
\end{minipage} \\
\midrule\noalign{}
\endhead
\bottomrule\noalign{}
\endlastfoot
\emph{10\% budget} & & & & \\
entropy (BLT's rule) & 10.6\% & 16\% & 19.0\% & 51.4\% \\
\texttt{=} rule (hand-written) & 10.6\% & 100\% & 51.8\% & 60.5\% \\
syntax rule: \texttt{=} and the answer marker & 10.6\% & 100\% & 51.5\%
& 61.2\% \\
boundary dependence & 10.0\% & 100\% & 67.8\% & 4.8\% \\
entropy + boundary dependence & 10.5\% & 100\% & 67.1\% & 62.9\% \\
SpaceByte-style (entropy after non-alphanumeric) & 10.6\% & 22\% &
22.2\% & 44.9\% \\
novelty table & 10.3\% & 100\% & 56.5\% & 19.4\% \\
novelty table + answer marker & 10.4\% & -- & 54.8\% & 65.6\% \\
corpus-fitted table & 10.0\% & 0\% & 10.2\% & 8.2\% \\
\emph{15\% budget} & & & & \\
entropy & 15.7\% & 48\% & 47.0\% & 65.0\% \\
\texttt{=} rule & 15.7\% & 100\% & 66.1\% & 67.0\% \\
entropy + boundary dependence & 15.7\% & 100\% & 76.5\% & 67.3\% \\
SpaceByte-style & 15.5\% & 68\% & 61.1\% & 66.0\% \\
novelty table & 15.4\% & 100\% & 66.5\% & 36.4\% \\
\end{longtable}
}
}

\textbf{What the table is.} The table is a cheap, learned detector of
cues that tend to precede new values in its fitting text. In GSM8K these
are \texttt{=} and \texttt{\$}, the same cues that keep entropy's signal
under the threshold: it covers every prose value after \texttt{\$} and
none after a plain space, and 4.9\% of the values of templated text. It
does not detect every fixed cue. In execution traces the same
\texttt{-\textgreater{}} precedes computed and copied values alike, so
on average it does not precede new content, and the table starts a patch
at neither (Appendix C). In training, what goes with accuracy is its
coverage of value starts (Table H1), and a hand-written rule shows no
detectable difference from it where both were trained (§4). A detector
of a different kind, a linear probe on BLT-1B's patcher state trained
without labels to predict ``a number not seen earlier in the text'',
also finds values in prose: forcing its top 3\% of positions gains +16.7
points over entropy on computed values (+12.7 to +20.9; the 711 prose
values above). That gain sits mostly on values after \texttt{\$},
shrinks as the budget grows (+14.4, +6.0 and +0.9 points at 10\%, 15\%
and 20\% on the 915 later problems), and costs final answers in prose
(-4.5 points, -6.4 to -2.6). It was tested at inference only, on GSM8K,
with prose from one writer (Appendix A). Across six domains at
inference, no rule we tested, the table included, beat entropy on exact
match of whole words and numbers in more than one domain (Appendix C).

\section{6. Budget and objective, at
inference}\label{budget-and-objective-at-inference}

Everything in this section is measured at inference on BLT-1B, a model
trained under entropy patching at about a quarter of bytes; only its
layout changes. These comparisons favour entropy's native layout on bits
per byte and say nothing about models trained under the other layouts;
the training-side versions were designed and not run (Appendix F).
Unless stated, the results use the 294 problems of Table 2, decision
rules fixed before each run, and paired problem-bootstrap intervals with
1,000 resamples (2,000 for the bits-per-byte comparisons computed for
this version: the \texttt{=} rule and the table against entropy, and the
splits into question and solution).

\textbf{The budget, by objective.} We sweep the patch budget (Table 3,
Figure 2) and ask how many patch starts each rule needs for the same
result. To reach the table's accuracy on computed results at 10\%
(56.5\%, at a 10.3\% patch rate), entropy needs about 1.75 times the
table's patch rate (problem-bootstrap interval 1.66 to 1.84; by the rule
fixed before the run, which asked for a lower bound of at least 1.5,
this holds). That interval leaves out the error of interpolating
linearly between entropy's measured points at 15.7\% and 21.0\% of
bytes, which alone bound the ratio between 1.52 and 2.04. Reported
without a verdict: to reach the \texttt{=} rule's computed-result
accuracy, entropy needs 1.59 times its rate (1.51 to 1.67), and to reach
the final-answer accuracy of the table with an answer marker, 1.57 times
(1.41 to 1.83). For whole-text bits per byte the trade runs the other
way: within the sweep the table never reaches entropy's 0.867 at 10\%
(it gets 0.944 at 15.4\% of bytes), so it needs more than 1.45 times
entropy's patch rate.

\textbf{What each start buys for perplexity.} On 150 of the problems we
measured the value of each patch start: the starts are split at random
into 5 groups and each group is removed in turn, and a start's value is
the rise in bits over the patch it started, so it includes interactions
within its group. 23.1\% of entropy's starts buy no bits (22.2 to 24.1),
against 43.2\% of the table's (41.8 to 44.7), and entropy's starts are
worth more on average (+1.97 bits against +0.87). Split by position
(after the data), most of the table's starts that buy nothing fall in
the question text (38\%) or inside words (23\%); its starts at spaces
and line breaks buy nothing most often (67\% and 82\% of them) but are
only 11\% of its starts.

\textbf{Both objectives at once.} Keeping entropy and adding a start at
each value (the \texttt{=} rule) improves both at 10\%: whole-text bits
per byte 0.857 against entropy's 0.867 (-0.010, -0.013 to -0.007; +0.011
on the question and -0.029 on the solution), and 51.8\% of computed
results against 19.0\%. A label-free version gets the answer gain with
no detectable whole-text difference in bits per byte. Forcing patch
starts at the positions whose table value is in the top 3\% of training
bytes, and filling the rest of the budget by entropy (the label-free
hybrid), gains +31.9 points on computed results over entropy at 10\%
(+27.7 to +35.7) and +12.6 on final answers (+8.2 to +17.7), with no
detectable difference in whole-text bits per byte (+0.005, -0.001 to
+0.010; HOLDS by the rule fixed before the run, which allowed an upper
bound of at most +0.02). That whole-text tie hides a shift between
regions: +0.032 bits per byte on the question (+0.027 to +0.038) and
-0.021 on the solution (-0.030 to -0.012). Against the \texttt{=} rule
the hybrid is tied on answers (computed results -0.9 points, -4.0 to
+2.3; final answers +3.4, -0.3 to +7.5) and costs 0.015 bits per byte
more (+0.009 to +0.020). With the top 1.5\% its whole-text bits per byte
are below entropy's (-0.006, -0.010 to -0.002) for a similar gain on
computed results (+31.2, +27.4 to +35.0); with the top 5\% the forced
starts cost bits (+0.061, +0.054 to +0.068).

\textbf{Tight budgets only.} At about BLT-1B's own rate (26.2\% of
bytes), entropy gets 77.8\% of computed results, and the hybrid's gain
over entropy falls from +31.9 points at 10\% to +20.9 at 15\% (+17.0 to
+24.4).

\textbf{Two cautions.} First, BLT-1B was trained under entropy patching,
which favours entropy's layout on bits per byte: at inference the table
alone costs +0.175 bits per byte (+0.164 to +0.186; §5), against +0.031
when trained from scratch at 50M (§4). The answers comparison therefore
runs against that home advantage, and the bits comparison with it.
Second, entropy plus boundary dependence, which gets more computed
results at 10\% than any layout in these curves (67.1\%, Table 2), is
not in them, so its bits per byte are unknown.

\needspace{27\baselineskip}

\textbf{Table 3.} BLT-1B at inference (a model trained under entropy
patching; only the layout changes): patch rate on the test text,
teacher-forced exact match (\%) on computed results and final answers,
and whole-text bits per byte, by layout and budget (294 problems). The
\texttt{=} rule forces a start at the first digit of each labelled
computed value (§5); the hybrids force the table's top q of positions
and fill the rest of the budget by entropy. Intervals for the
differences are in the text.

{\def\LTcaptype{none} 
{\footnotesize\setlength{\tabcolsep}{3pt}
\begin{longtable}[]{@{}
  >{\raggedright\arraybackslash}p{(\linewidth - 8\tabcolsep) * \real{0.5970}}
  >{\raggedright\arraybackslash}p{(\linewidth - 8\tabcolsep) * \real{0.0746}}
  >{\raggedright\arraybackslash}p{(\linewidth - 8\tabcolsep) * \real{0.1194}}
  >{\raggedright\arraybackslash}p{(\linewidth - 8\tabcolsep) * \real{0.0746}}
  >{\raggedright\arraybackslash}p{(\linewidth - 8\tabcolsep) * \real{0.1343}}@{}}
\toprule\noalign{}
\begin{minipage}[b]{\linewidth}\raggedright
layout
\end{minipage} & \begin{minipage}[b]{\linewidth}\raggedright
rate
\end{minipage} & \begin{minipage}[b]{\linewidth}\raggedright
computed
\end{minipage} & \begin{minipage}[b]{\linewidth}\raggedright
final
\end{minipage} & \begin{minipage}[b]{\linewidth}\raggedright
bits/byte
\end{minipage} \\
\midrule\noalign{}
\endhead
\bottomrule\noalign{}
\endlastfoot
entropy & 5.5\% & 7.8\% & 27.6\% & 1.044 \\
entropy & 10.6\% & 19.0\% & 51.4\% & 0.867 \\
entropy & 15.7\% & 47.0\% & 65.0\% & 0.764 \\
entropy & 21.0\% & 68.5\% & 71.4\% & 0.706 \\
entropy & 26.2\% & 77.8\% & 73.1\% & 0.684 \\
entropy & 31.3\% & 79.1\% & 72.8\% & 0.675 \\
novelty table & 5.1\% & 27.4\% & 13.6\% & 1.149 \\
novelty table & 7.7\% & 41.6\% & 18.4\% & 1.099 \\
novelty table & 10.3\% & 56.5\% & 19.4\% & 1.042 \\
novelty table & 15.4\% & 66.5\% & 36.4\% & 0.944 \\
novelty table + answer marker & 10.4\% & 54.8\% & 65.6\% & 1.043 \\
entropy + a start at each value (\texttt{=} rule) & 8.0\% & 36.7\% &
55.1\% & 0.939 \\
entropy + a start at each value (\texttt{=} rule) & 10.6\% & 51.8\% &
60.5\% & 0.857 \\
entropy + a start at each value (\texttt{=} rule) & 15.7\% & 66.1\% &
67.0\% & 0.757 \\
entropy + table's top 1.5\% (label-free hybrid) & 10.6\% & 50.1\% &
63.9\% & 0.861 \\
entropy + table's top 3\% (label-free hybrid) & 10.5\% & 50.9\% & 63.9\%
& 0.872 \\
entropy + table's top 5\% (label-free hybrid) & 10.5\% & 51.0\% & 65.0\%
& 0.928 \\
entropy + table's top 3\% (label-free hybrid) & 15.6\% & 67.8\% & 68.4\%
& 0.762 \\
\end{longtable}
}
}

\begin{figure}
\centering
\includegraphics[width=1\linewidth,height=\textheight,keepaspectratio,alt={BLT-1B
at inference: teacher-forced exact match on computed results (left) and
final answers (right) against whole-text bits per byte, by layout and
budget (the layouts of Table 3; labels give the budget). Up and to the
left is better on both objectives. The legend's ``entropy + start after
`='\,'' is the = rule, which forces a start at the first digit of each
labelled computed value (§5). Drawn by
paper2/v2/scripts/fig\_tradeoff.py from blt\_waste.json and
blt\_hybrid.json.}]{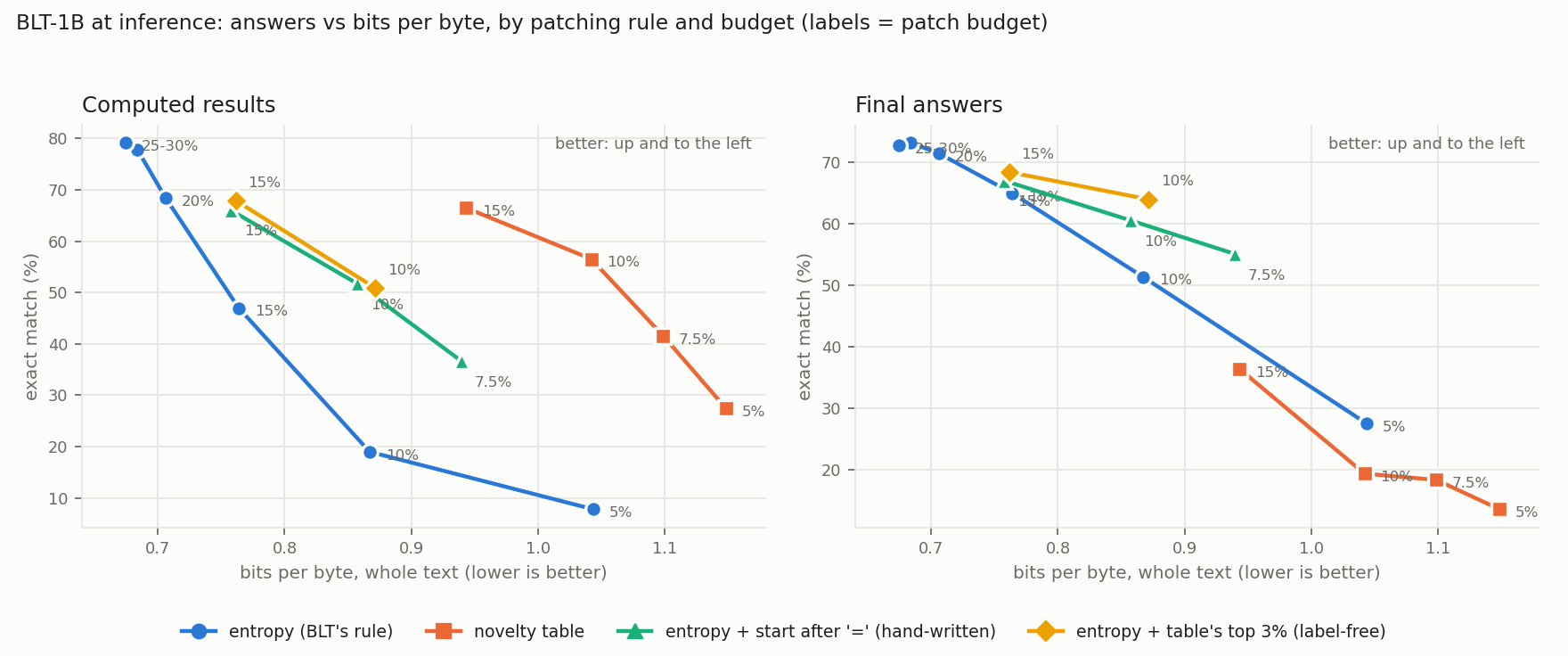}
\caption{BLT-1B at inference: teacher-forced exact match on computed
results (left) and final answers (right) against whole-text bits per
byte, by layout and budget (the layouts of Table 3; labels give the
budget). Up and to the left is better on both objectives. The legend's
``entropy + start after `='\,'' is the \protect\texttt{=} rule, which
forces a start at the first digit of each labelled computed value (§5).
Drawn by \protect\texttt{paper2/v2/scripts/fig\_tradeoff.py} from
\protect\texttt{blt\_waste.json} and \protect\texttt{blt\_hybrid.json}.}
\end{figure}

\textbf{What a patch start buys, by kind of unit.} Within the same
BLT-1B texts at 10\%, with a random four-digit ticket number inserted
twice (new and unpredictable, then a pure copy), we measured the value
of a patch start at each unit: bits with the start removed minus bits
with it added, the rest of the entropy layout fixed (Table 4; 293
problems). The comparisons and their decision rules were fixed after a
six-problem smoke test, while the full run was under way and before any
of its results was read. A start buys 3.13 bits at computed results and
1.93 at final answers; it costs at the random ticket (-0.72) and at its
pure copy (-1.33; exact copying falls from 70.3\% without the start to
26.3\% with it). Copied numbers in the solution gain 1.21 bits, more
than a pure copy; a reading formed after the data is that choosing which
earlier number comes next takes work. Among new words, those an
independent model (Qwen2.5-1.5B \citep{yang2024qwen25}) finds easy gain
1.16 bits more than the hardest third (+0.89 to +1.44; 2,000 resamples),
though the hardest third still gains. By the rules fixed in advance,
``copies gain little'' was inconclusive (copied numbers gain 39\% as
much as computed results) and the learnability comparison held.

\needspace{13\baselineskip}

\textbf{Table 4.} Value of a patch start by kind of unit (BLT-1B, 10\%,
293 problems; bits per unit, start removed minus start added;
problem-level 95\% intervals, 1,000 resamples).

{\def\LTcaptype{none} 
{\footnotesize\setlength{\tabcolsep}{3pt}
\begin{longtable}[]{@{}llll@{}}
\toprule\noalign{}
unit & n & gain (bits) & interval \\
\midrule\noalign{}
\endhead
\bottomrule\noalign{}
\endlastfoot
computed result & 792 & +3.13 & {[}+2.92, +3.33{]} \\
final answer & 293 & +1.93 & {[}+1.77, +2.10{]} \\
new word & 1,986 & +1.73 & {[}+1.61, +1.85{]} \\
copied number & 2,336 & +1.21 & {[}+1.13, +1.29{]} \\
repeated word & 7,018 & +0.59 & {[}+0.54, +0.63{]} \\
random ticket number & 293 & -0.72 & {[}-0.86, -0.59{]} \\
copy of the ticket & 293 & -1.33 & {[}-1.44, -1.24{]} \\
\end{longtable}
}
}

\section{7. Other results, in brief}\label{other-results-in-brief}

\textbf{A label-free detector (Appendices A and E).} Appendix A gives
the probe of §5 in full, with boundary dependence beside it. Within
numbers the probe separates computed values from the other numbers by
the phrase before them (AUC 0.77, 0.74 to 0.80). On the 915 later
problems no layout tested keeps the whole prose gain without a
final-answer loss; a fused ranking chosen on training problems keeps a
smaller gain (+11.1 points, +9.2 to +13.0) with no detectable
final-answer difference (-0.7, -2.2 to +0.8). All of this is at
inference on BLT-1B, on GSM8K, with prose from one LLM writer. A
one-seed 12M model trained under the probe hybrid (Appendix E) is an
exploratory pilot, not a test; the three-seed test was not run.

\textbf{Starts inside a verbatim copy (Appendix B).} Two follow-ups of
Table 4's last row, with decision rules filed before scoring, used GSM8K
training problems new to this work by index, in the same template. At
BLT-1B's default rate, removing entropy's starts on a copied ID raises
exact copies from 13.5\% to 88.3\% (+74.8 points, +71.3 to +78.2; 600
texts, 512-byte windows restored), and a veto gated by the patcher's own
prediction (L8) recovers +73.7 of those points (+70.2 to +77.2), and
+13.8 at 10\% (+10.8 to +17.0), with its costs inside the filed bars
read on point estimates. A veto computed from bytes alone (SIV+PO) was
better than L8 inside the windows in one cell, 4-digit tickets at the
default rate (L8 minus SIV+PO -2.0 points, -3.3 to -0.8; tied in the
other four comparisons), and worse under the conversion as released
(+6.7, +4.0 to +9.5), which lacks the windows and where, in texts of 700
tokens or more, removing entropy's starts on the copy loses every copy.
Stopped at token 620, the byte veto then beat L8 under the released
conversion at the default rate (L8 minus the byte veto: -3.17 points of
exact copies, -5.00 to -1.33), on exact copies only (bits per copy show
no detectable difference, -0.014, -0.074 to +0.044), at that rate only,
with about 9\% of replications tying, a cutoff set after earlier results
had been seen that sits on these texts' own optimum, and no claim about
costs. All of this rests on synthetic IDs in one template, one model and
teacher forcing.

\textbf{Outside math (Appendix C).} Trained from scratch at 12M on
prose, Python code and execution traces (2 seeds per rule, all on T4s in
fp16), the table and entropy show no detectable difference on any
measure. At inference on BLT-1B, three tests with rules fixed beforehand
asked whether any rule predicts where an extra 3\% of patch starts pays,
in six domains (GSM8K, MATH algebra, Python, Wikipedia, clinical
vignettes, execution traces). On exact match of whole words and numbers,
entropy plus the table beat entropy only in code, maps of measured gains
and probes on the patcher state fitted on other domains beat it in at
most one domain, and no rule beat it in more than one. A structure-only
rule that adds starts at word and number starts beat entropy on bits in
five of six domains (exploratory, no verdict set; consistent with
SpaceByte's word boundaries); on whole words and numbers it equals
entropy by construction.

\textbf{What did not work (Appendix D).} Classic compressors used
directly as patch rules (document-level LZ, PPM-C) are below entropy on
BLT-1B. Value uncertainty finds computed results at inference but fails
as a training rule at 12M (in a comparison across devices). A 12M model
trained under the table loses nearly all final answers when its layout
is switched to entropy at inference.

\textbf{Work that was not run (Appendix F).} Designed with decision
rules and not run when the line of work closed: Study A beyond its
interim look; Study B, BLT-1B fully fine-tuned under the table; the
hand-written rule at 50M with the table's recipe; a 12M training budget
curve with a same-device anchor; and both hybrids trained from scratch.
No results exist for any of these.

\section{8. Limitations}\label{limitations}

\textbf{Scale and recipe.} Models trained from scratch reach 50M
parameters, and the methods we compare against run at 1B and above. At
1M and 12M the table ran on T4 GPUs in fp16 and the baselines on an A40
in bf16, unclipped; at 50M every run in Table 1 was clipped, and every
50M run used the A40. Study A (145M) was stopped after an interim look
at one seed.

\textbf{Evaluation.} Every accuracy is teacher-forced exact match on
answer spans; end to end, the 1M-50M models trained from scratch solve
at most 2 of 100 GSM8K problems. Final answers mostly copy the last
computed result. The models see each training solution about 1.4 to 16
times. Three seeds per configuration (two outside math) leave many
differences unresolved.

\textbf{BLT-1B.} Every BLT-1B result is at inference, on one model
trained under entropy patching, through a community conversion with an
unmerged window fix, at budgets below the rate it was trained with.

\textbf{The cue.} The ceiling was measured on BLT-1B's 14-layer patcher
only; whether it explains why the one-layer patchers of §4 cover 0.7\%
of computed results was not tested. The 50M control with the
hand-written rule was not run, so whether the table does more than a cue
rule in training is open at the one size where it beats entropy on both
targets. No word-boundary, fixed-stride, learned-chunker or line-start
rule was trained from scratch, and a table fitted on OpenWebMath alone
was not tested.

\textbf{Domain.} The positive results concern computed values in written
math with fixed cues, at tight budgets. In six domains, no rule we
tested beat entropy on exact match of whole words and numbers in more
than one domain.

\textbf{Detector.} Inference only, GSM8K only, prose from one LLM
writer; most of its prose gain sits on values after \texttt{\$}, a cue
the rewrite kept.

\textbf{Copies.} Synthetic IDs in one template, one model,
teacher-forced; the comparison with the cutoff rule holds on exact
copies only, at one patch rate and one placement, chosen after earlier
results were seen.

\textbf{Statistics.} Many intervals are reported without a correction
for multiple comparisons, and analyses not planned before their data are
marked and exploratory. The decision-rule files are public in
\texttt{paper2/v2/rules/}, but the commits that date them are in a
private repository.

\section{9. Conclusion}\label{conclusion}

Under tight patch budgets, BLT-1B's entropy patching misses computed
values in worked math because the cue before a value fixes the next
byte's type, which caps the patcher's entropy near the threshold exactly
where the value begins. A two-byte table, motivated by LZ novelty,
learns cues that tend to precede new values in its fitting text, here
\texttt{=} and \texttt{\$}; in execution traces, where the same cue
precedes copies, it starts a patch at no value. Trained from scratch, it
beats entropy at 50M on teacher-forced answer spans, but where both were
trained (1M and 12M) a hand-written rule with four markers shows no
detectable difference from it. At inference on BLT-1B, the table falls
below entropy on prose without the \texttt{=} cue; on solutions written
with \texttt{=}, adding starts at the table's strongest positions to
entropy's layout raises exact match on computed values from 19.0\% to
50.9\% at a 10\% budget, with no detectable difference in whole-text
bits per byte, though bits per byte move from the solution to the
question; and a label-free probe of the patcher state finds values in
prose at tight budgets, mostly after \texttt{\$}. In six domains, no
rule we tested beat entropy on exact match of whole words and numbers in
more than one domain. This line of work is closed; the training-side
tests that would show whether these layouts help a model trained with
them were designed and not run.

\section{Changes in version 2}\label{changes-in-version-2}

This version replaces version 1 (2026-10-15) of arXiv:2610.22100. After
version 1 the work was revised in private for two weeks; on 2026-11-01
an independent review found that version 1 overclaims, and the author
closed this line of work. No experiments were run after that. Version 2
is a corrected and complete record of the work's final state; no venue
is intended. Each item is marked \emph{withdrawn}, \emph{corrected}
(replaced by the stated fact) or \emph{qualified} (it stands within the
stated limits).

\begin{itemize}
\tightlist
\item
  \textbf{Title and framing. Withdrawn.} Version 1's title, ``Patch
  where the text is new: a compression view of where byte-level models
  should spend compute'', is withdrawn: the table and entropy select new
  content at similar rates (65.3\% and 64.2\% of their selections; the
  table slightly more, Appendix H), the true LZ target covers fewer
  values than entropy where no fixed cue precedes them, and on prose the
  table falls below entropy (§5). The paper is framed by the cue ceiling
  (§5); LZ novelty is kept only as the motivation of the table.
\item
  \textbf{What stands.} Version 1's training runs and Table 1 means, now
  with intervals and the disclosures of §3, and the table's 50M gains
  over entropy and dependence on one device (§4).
\item
  \textbf{Claims of version 1} (version 1's words in quotation marks).

  \begin{itemize}
  \tightlist
  \item
    \emph{Corrected.} The corpus-fitted table ``keeps the gain'': it
    keeps most of it (final answers -3.1 points, -5.9 to -0.4), its
    corpus contains the curated solutions, and on BLT-1B's format it
    covers none of the computed results (§4). \emph{Withdrawn:} ``the
    rule does not depend on curated math examples'' and its ``better
    bits per byte''.
  \item
    \emph{Withdrawn.} ``At 12M it has the best bits per byte of all
    rules (1.619)'' (the corpus-fitted table): the comparison crosses
    devices (the table arms on T4s, the other rules on the A40).
  \item
    \emph{Qualified.} ``From 47\% to 72\% at 12M'': the scores are
    teacher-forced, the comparison crosses devices, and the table's mean
    is 71.5\% (§3, §4).
  \item
    \emph{Qualified.} ``At a matched \textasciitilde11\% patch rate'':
    the rates were matched on training bytes (10\%; 13\% for the
    corpus-fitted table) and range from 10.9\% to 12.1\% on evaluation
    text (§3).
  \item
    \emph{Withdrawn.} ``The effect grows with model size'': the effect
    is present at 12M on final answers (across devices) and at 50M on
    both targets, with recipe and device changing between sizes (§4).
  \item
    \emph{Corrected.} Clipping ``checked to be neutral for the
    baselines'': it lowered entropy's bits per byte at 50M (§3).
  \item
    \emph{Withdrawn.} ``Novelty lowers bits per byte on prose'',
    ``novelty is better on copied values'' (traces) and, on code, ``at
    an equal training budget novelty is better'': no difference is
    distinguishable at two seeds, and the higher-budget code pair is not
    rate-matched (Appendix C).
  \item
    \emph{Withdrawn.} ``An analysis of where boundaries fall explains
    the pattern'': the analysis is descriptive and partly circular
    (Appendix H).
  \item
    \emph{Qualified.} ``Computed results have ordinary surprisal for the
    patcher's model'': this holds for the one-layer patcher of the
    models trained from scratch (3.34 bits at computed-result starts
    against 3.12 over all bytes; \texttt{results/enrichment.json} and
    the lab notebook, \texttt{notes.md}), not for BLT-1B's patcher,
    whose entropy at computed-result starts is above average (2.90
    against 1.11 bits) yet under a tight budget's threshold (§5).
    \emph{Withdrawn:} ``stop treating surprisal as the target'': the
    high-enrichment hand-written rule shows no detectable difference
    from the table at 12M (§4), and in six domains no rule we tested
    beat entropy on exact match of whole words and numbers in more than
    one (Appendix C).
  \item
    \emph{Withdrawn.} ``Novelty works where new content begins at
    segment starts close to the computation'' and ``A boundary at the
    value is not what helps'': on GSM8K the table detects the cue before
    values, and on BLT-1B a start at a computed value buys 3.13 bits
    (§6); the segment-start reading is untested (Appendix H).
  \item
    \emph{Corrected.} The H-Net measurement attributed to
    \citet{vera2026easy} is not in that paper; it is reported with its
    caveats in Appendix G. \emph{Withdrawn:} the statement that learned
    chunkers find such structure more generally, and the H-Net numbers
    on Python.
  \item
    \emph{Corrected.} ``The BLT paper does not describe it'' (a lookup
    patcher): BLT's §4.2 notes that a small-context entropy model can be
    encoded as a lookup table, and its Fig. 3(f) uses a two-byte-context
    entropy model (§2). \emph{Withdrawn:} ``Our results are consistent
    with ByteFlow's direction''.
  \item
    \emph{Corrected.} Value uncertainty ``fails when used for training
    at every size'': it fails at 12M (across devices); at 50M the
    difference is not distinguishable, and at 1M it is near the floor,
    as is entropy (Appendix D). \emph{Withdrawn:} ``with the best bits
    per byte of all rules'': at 12M its bits per byte (1.678) are above
    entropy's (1.657), and at 50M the difference is not distinguishable
    (-0.006, -0.033 to +0.021).
  \item
    \emph{Corrected.} The table plus an answer marker as ``the best
    label-free layout we found without retraining'': entropy plus
    boundary dependence, also label-free, scores higher on computed
    results (§5).
  \item
    \emph{Corrected.} The rule switch ``falls from 73.6\%'': the 73.6\%
    is the mean of two separate T4 re-runs, not of Table 1's seeds
    (Appendix D).
  \item
    \emph{Corrected.} ``It costs nothing at inference'': the lookup
    itself is cheap, but the layout costs bits per byte in training (§4)
    and, alone on BLT-1B, final answers and bits per byte (§5).
  \item
    \emph{Withdrawn.} Combining novelty with entropy ``was worse than
    either alone'': the rule was misdescribed, and the result is neither
    in \citet{vera2026easy} nor in any public results file; withdrawn
    with the sentence on learned predictors of boundary dependence,
    whose numbers are in no public results file.
  \item
    \emph{Corrected.} ``Fitted in seconds on raw text with no labels'':
    no labels are used, but the table is fitted on curated solutions in
    the scored format (§3).
  \end{itemize}
\item
  \textbf{New disclosures} (§3). Devices and precisions, teacher forcing
  and the end-to-end floor, final answers as copies, how often the
  models see each training solution, and the third-party BLT-1B
  conversion with budgets outside its training.
\item
  \textbf{New results.} Intervals for differences between rules; the cue
  ceiling, the prose rewrite and cue rules at inference (§5); the budget
  by objective and the label-free hybrid (§6); the detector, the copy
  tests, the six-domain tests, other negative results, a one-seed
  training pilot and the H-Net check (Appendices A-E and G); Study A's
  interim look (Appendix F).
\item
  \textbf{Not run.} Planned work that was not run is listed in Appendix
  F, without results.
\item
  \textbf{Numbers.} The training runs behind Table 1 are those of
  version 1; the 1M dependence means now leave out the seed that never
  learned, as every interval does. The table's description now follows
  the code (a 2 MB fitting sample and a window of the previous 127
  bytes; version 1 said 4 MB and 128). Version 1's code exact-match rows
  used token-level standard errors; they now carry seed-level intervals,
  and they are stated to come from separately trained runs
  (\texttt{results/generic/code\_acc/}), not from the runs of Table C1
  (Appendix C). Version 1's Figure 2 (entropy's bits per byte on code
  over four budgets) is dropped with the equal-budget claim it
  supported. The table's share of boundaries at line starts in math is
  9\% (version 1: 10\%; Appendix H). Differences between rules carry
  seed-level or paired problem-level 95\% intervals, except where the
  text says that none was computed or that the comparison is
  descriptive.
\end{itemize}

\section{Code and data}\label{code-and-data}

Code, configurations, the corpus recipe (SHA-256), per-run results, the
dated decision-rule files, the preregistration with amendments A-1 to
A-3 (A-4 is summarized and A-5 and A-6 are recorded in
\texttt{paper2/v2/rules/STUDY\_A\_INTERIM\_RULES.md}), the prose
rewrites, the copy-test texts, and the scripts and results behind every
table, figure and number are at
\url{https://github.com/nicoveraz/segresearch}, release v2.1.0
(doi:10.5281/zenodo.24023934). The models trained from scratch are in
\texttt{results/scale/} and \texttt{results/generic/}; the results
specific to this version, with their scripts, rules and provenance, are
in \texttt{paper2/v2/} (see \texttt{paper2/v2/README.md}); and
\texttt{tests/test\_paper2\_numbers.py} checks the paper's result
numbers against those files. Some results cannot be recomputed from the
release alone, because their inputs (the 1.9 GB training corpus and
caches derived from it) are not shipped or because they need BLT-1B; the
README lists them. The commits that date the decision rules are in the
author's private working repository, so their timing cannot be checked
publicly.

\textbf{Compute.} BLT-1B inference, the small-patcher variants of
Appendix A and the pilot of Appendix E on an Apple laptop GPU (MPS); the
1M and 12M table and value-uncertainty runs, all runs outside math and
Study A's 145M runs on Kaggle T4 GPUs (fp16); all other training on
rented NVIDIA A40 GPUs (bf16).

\textbf{Prose rewrites.} The GSM8K solutions (MIT licence) used in §5
and Appendix A were rewritten as prose by Claude Opus 5.5 (model ID
\texttt{claude-opus-5-5}) with one fixed prompt, through Anthropic's
batch API: 1,382 training and 275 test problems kept for the first test
(221 test problems with 711 computed values) and 915 of 1,019 new test
problems for the later ones, for about \$11 in all. Automatic checks
kept only rewrites whose marked values equal the original's computed
values in order, with no \texttt{=} and no operator between numbers;
nothing checks that the prose explains the solution correctly. A second
writer was planned and not used. The data card is in
\texttt{paper2/v2/README.md}.

\textbf{Use of AI.} The experiments, analyses and this revision were
carried out with an AI coding agent (Claude Code) working under the
author's direction; the prose rewrites were written by a language model,
as described above. \textbf{Funding:} none. \textbf{Competing
interests:} none.

\bibliography{refs}

\section{Appendix A. A label-free detector on BLT-1B's patcher
state}\label{appendix-a.-a-label-free-detector-on-blt-1bs-patcher-state}

Every result here is at inference on BLT-1B, at a 10\% budget unless
stated, teacher-forced, on GSM8K problems in two writings:
\emph{equation text} (the solutions without calculator annotations,
values after \texttt{=}) and the prose rewrites of §5, from one LLM
writer. Each test had decision rules filed before its run
(\texttt{paper2/v2/rules/}, R10 to R22); as a series they are
exploratory, since each was designed after the earlier results. The
first tests use the 275 test problems of §5 (221 of them with computed
values; 711 values), the later ones 915 test problems that none of the
earlier detector tests had scored (2,448 values; their original writings
are among the 1,292 solutions of §5). Intervals: paired problem
bootstrap, 2,000 resamples.

\textbf{The probe.} At each candidate position (a byte after a
non-alphanumeric byte, where a patch may start), a logistic probe reads
BLT-1B's patcher state before the candidate byte (the last layer's
normed hidden state, reduced to 32 principal components). Simple probes
on a language model's hidden states can recover arithmetic results and
flag arithmetic errors \citep{sun2025probing}; here a probe reads the
patcher's state to decide where a patch starts. A layout forces starts
at the candidates whose score is in the top 3\% of training positions
and fills the rest of the budget by entropy, as the hybrid of §6 does
with the table. Trained on the rewrite's value markers (supervised), the
probe detects computed-value starts in prose with AUC 0.97 (0.96 to
0.97) when trained on prose and 0.72 (0.69 to 0.75) when trained on
equation text, against 0.66 for entropy, and forcing its top 3\% gains
+14.5 and +8.0 points on computed values over entropy (+10.9 to +18.2;
+4.5 to +11.8). Trained without labels, on ``a number not seen earlier
in the text'' (auto-labels; against the true computed values in prose,
precision 0.64 and recall 0.89), it gains +16.7 points in prose (+12.7
to +20.9), slightly above the supervised probe refitted in the same run
(+14.9; difference +1.8, +0.1 to +3.8); trained on equation text and
applied to prose, +14.5 (+10.8 to +18.3); in equation text, +54.6 (+50.8
to +58.5). In prose it lowers bits per byte of the solution (-0.012,
-0.017 to -0.006) but loses final answers (57.5\% against 62.2\%).

\textbf{What it reads.} Within numbers, the auto-labelled probe
separates computed values from the other numbers by the phrase before
them (AUC 0.77, 0.74 to 0.80; 711 computed values among 3,085 numbers in
the prose test), where a probe for ``any number'' gets 0.59. At a 3\%
share it forces more starts than there are numbers (5,838 against 3,085
numbers in prose, 59\% of them on non-numbers), so the any-number probe
gets most of the gain: the auto-labelled probe is ahead of it by +2.8
points in prose (+0.4 to +5.3) and behind it by 3.5 in equation text
(-5.5 to -1.6). Around a number (descriptive), the patcher's entropy is
alike for computed values and copies at the first digit (AUC 0.56) and
separates them from the second digit on (0.84 at the second digit), and
the probe's score rises at the decision point (AUC 0.84 there, 0.61 one
byte before). A start at a number's second digit helps the rest of the
number but whole computed values only a little (+4.1 points, +2.2 to
+6.1, against +11.4 for a start at the first digit, +8.7 to +14.2).

\textbf{Where the gain sits, and its costs} (Table A1; the 915 later
problems). The 3\% layout gains +14.4 points in prose (+12.2 to +16.6).
Most of it sits on values after \texttt{\$}, a cue the rewrite kept:
+47.1 points there (+42.5 to +51.8; 575 values) against +4.4 after a
plain space (+2.6 to +6.2; 1,873 values); this split was seen before the
rule was written. The gain shrinks as the budget grows (+6.0 at 15\%,
+4.4 to +7.6; +0.94 at 20\%, +0.04 to +1.82), and it beats a random 3\%
of candidates by +12.9 (+10.8 to +15.1). In prose the layout loses final
answers (-4.5 points, -6.4 to -2.6) and costs whole-text bits per byte
(+0.0085, +0.0063 to +0.0107; +0.048 on the question, -0.010 on the
solution). About half of the final-answer loss comes from the 67
problems where its forced starts displace entropy's start at the final
answer (-2.2, -3.3 to -1.3); keeping entropy's final-line starts
recovers +2.0 (+0.9 to +3.1) but still loses 2.5 (-4.4 to -0.8). Forcing
only as many starts as there are numbers (1.53\% of prose positions)
keeps +12.3 on computed values (+10.2 to +14.5) and still loses 1.9 on
final answers (-3.5 to -0.1). A fused ranking chosen on training
problems (entropy plus twice the probe's probability, with no forced
share) keeps +11.1 (+9.2 to +13.0) with no detectable final-answer
difference (-0.7, -2.2 to +0.8) and slightly lower whole-text bits per
byte (-0.003, -0.004 to -0.001). A fused layout that needs no probe,
with value uncertainty in its place, gains +36.3 on prose values after
\texttt{\$} (+31.8 to +41.2) but loses 5.0 points of final answers (-7.2
to -3.0). No layout tested keeps the whole gain without a final-answer
loss. In equation text the 3\% layout gains +54.3 over entropy and +20.0
over the \texttt{=} rule (+17.9 to +22.0), raises final answers (+14.6,
+11.8 to +17.6) and lowers whole-text bits per byte (-0.013, -0.017 to
-0.010). Boundary dependence \citep{vera2026easy} is close behind the
probe on computed values in both writings (2.6 points behind in prose,
+0.5 to +4.7; 2.0 in equation text, +0.2 to +3.7); in prose it also
raises final answers over entropy (+3.8, +1.1 to +6.4), at +0.096 bits
per byte on the whole text (+0.155 on the question), while in equation
text it drops them to 2.3\%.

\needspace{17\baselineskip}

\textbf{Table A1.} Detector layouts on the 915 later GSM8K test problems
(BLT-1B, 10\% budget, teacher-forced): exact match (\%) on computed
values and final answers in each writing, and whole-text bits per byte
minus entropy's in prose. Point estimates; intervals for the differences
are in the text.

{\def\LTcaptype{none} 
{\footnotesize\setlength{\tabcolsep}{3pt}
\begin{longtable}[]{@{}
  >{\raggedright\arraybackslash}p{(\linewidth - 10\tabcolsep) * \real{0.5063}}
  >{\raggedright\arraybackslash}p{(\linewidth - 10\tabcolsep) * \real{0.1013}}
  >{\raggedright\arraybackslash}p{(\linewidth - 10\tabcolsep) * \real{0.0759}}
  >{\raggedright\arraybackslash}p{(\linewidth - 10\tabcolsep) * \real{0.1139}}
  >{\raggedright\arraybackslash}p{(\linewidth - 10\tabcolsep) * \real{0.1013}}
  >{\raggedright\arraybackslash}p{(\linewidth - 10\tabcolsep) * \real{0.1013}}@{}}
\toprule\noalign{}
\begin{minipage}[b]{\linewidth}\raggedright
layout
\end{minipage} & \begin{minipage}[b]{\linewidth}\raggedright
prose: computed
\end{minipage} & \begin{minipage}[b]{\linewidth}\raggedright
prose: final
\end{minipage} & \begin{minipage}[b]{\linewidth}\raggedright
prose: bits/byte vs entropy
\end{minipage} & \begin{minipage}[b]{\linewidth}\raggedright
equation text: computed
\end{minipage} & \begin{minipage}[b]{\linewidth}\raggedright
equation text: final
\end{minipage} \\
\midrule\noalign{}
\endhead
\bottomrule\noalign{}
\endlastfoot
entropy & 21.7 & 59.0 & -- & 19.3 & 48.5 \\
entropy + a start at every value (labels) & 33.2 & 59.3 & -0.0074 & 56.2
& 61.9 \\
boundary dependence & 33.5 & 62.8 & +0.0960 & 71.7 & 2.3 \\
probe, top 3\% & 36.1 & 54.5 & +0.0085 & 73.6 & 63.2 \\
probe, top 3\%, entropy's final-line starts kept & 35.9 & 56.5 & +0.0098
& 73.2 & 64.7 \\
probe, as many starts as numbers & 34.0 & 57.2 & -- & 72.8 & 63.1 \\
fused ranking (entropy + 2 × probe) & 32.8 & 58.4 & -0.0027 & 59.6 &
62.0 \\
fused, value uncertainty instead of the probe & 29.0 & 54.0 & +0.0047 &
54.3 & 60.2 \\
random 3\% of candidates & 23.2 & 52.5 & +0.0171 & 33.6 & 52.7 \\
\end{longtable}
}
}

\textbf{Training the small patcher.} On the one-layer, width-32 patcher
of our models trained from scratch (4,000 updates of 32 windows of 128
bytes, about 16M bytes, one seed per variant, on the laptop GPU), adding
a head for new numbers, a look-ahead loss or both did not raise the
probe's AUC for computed values in prose (-0.008, -0.007 and -0.002;
each interval includes 0), while a larger patcher (width 64, 2 layers)
did (+0.022, +0.013 to +0.032). In equation text every change helped a
little, each less than the larger patcher. In training from scratch
there is only the one-seed pilot of Appendix E.

\section{Appendix B. Patch starts inside a verbatim
copy}\label{appendix-b.-patch-starts-inside-a-verbatim-copy}

Table 4's last row shows that, at 10\%, a patch start on a copied ticket
number costs bits and breaks exact copying. Two follow-ups, O6b and O6c
(the test names used in the rules files), tested this with decision
rules filed before any new text was scored
(\texttt{paper2/v2/rules/O6b\_COPY\_LATCH\_RULES.md} and
\texttt{O6c\_CUTOFF\_RULES.md}), each followed by an independent check
of the run and its reading (\texttt{paper2/v2/checks/}). Both are
teacher-forced on BLT-1B's base weights (\texttt{itazap/blt-1b-hf}), on
GSM8K training problems new to this work by index: not selected before
for scoring, threshold fitting or design (O6b from index 1,800, O6c from
index 6,253). They are among the solutions in the training corpus and
the table's fitting set, and about half of each test's pool was among
the problems drawn to adapt BLT-1B in \citet{vera2026easy} (47\% of
O6b's pool and 554 of the 1,220 problems of O6c's, when the draws are
replayed, as the rules files record), whose adapted weights are not used
here. The texts are written in Table 4's template:
\texttt{Ticket\ NNNN.} before the solution and the copy line
\texttt{Ticket\ NNNN\ closed.} before the final line. Development texts,
which include Table 4's problems, were used to design the vetoes;
exploratory runs on them first showed the hazard
(\texttt{blt\_copy\_gate.json}, \texttt{blt\_copy\_veto.json} and
\texttt{blt\_copy\_latch\_step*.json} in \texttt{paper2/v2/results/}).
Intervals: paired problem bootstrap, 2,000 resamples.

\textbf{Layouts.} \emph{Entropy} is BLT-1B's layout at its default
threshold (the default rate) or refitted to 10\%. \emph{No start on the
copy} is entropy without its starts on the copied ID, not refilled.
\emph{L8} vetoes starts from a copy's first byte to its end when, among
other conditions (the full definition is in the O6b rules file), the 8
bytes before it recur, the earlier string was unpredictable to the
patcher (at least 10 bits of summed entropy) and the patcher predicts
the copy's first byte; it reads the model. \emph{SIV+PO} (a
segment-initial veto plus a pointer-onset veto) vetoes starts from bytes
alone: where the line so far repeats the start of earlier lines that all
continued alike, or where a recurring context of 8 or more bytes points
to an ID-like string seen once. \emph{SIV+PO\textless620} is SIV+PO at
token positions before 620 only (token t is byte t-1; token 0 is the
beginning-of-sequence token). Under each veto, entropy's threshold is
refitted on 300 separate fit texts to entropy's rate on them, or to
10\%. A paired comparison gets a class only if the difference is at
least 2 points with its 95\% interval on that side of 0; otherwise it is
a tie, which is not an equivalence test.

\needspace{13\baselineskip}

\textbf{Table B1.} Exact match (\%) of the whole copied ID,
teacher-forced, by layout. O6b's first four rows restore BLT-1B's
512-byte windows; the last two use the conversion as released (no
windows). Entropy's patch rate on the test texts is in parentheses.

{\def\LTcaptype{none} 
{\footnotesize\setlength{\tabcolsep}{3pt}
\begin{longtable}[]{@{}
  >{\raggedright\arraybackslash}p{(\linewidth - 12\tabcolsep) * \real{0.5263}}
  >{\raggedright\arraybackslash}p{(\linewidth - 12\tabcolsep) * \real{0.0526}}
  >{\raggedright\arraybackslash}p{(\linewidth - 12\tabcolsep) * \real{0.0921}}
  >{\raggedright\arraybackslash}p{(\linewidth - 12\tabcolsep) * \real{0.0658}}
  >{\raggedright\arraybackslash}p{(\linewidth - 12\tabcolsep) * \real{0.0526}}
  >{\raggedright\arraybackslash}p{(\linewidth - 12\tabcolsep) * \real{0.0789}}
  >{\raggedright\arraybackslash}p{(\linewidth - 12\tabcolsep) * \real{0.1316}}@{}}
\toprule\noalign{}
\begin{minipage}[b]{\linewidth}\raggedright
test, text and rate
\end{minipage} & \begin{minipage}[b]{\linewidth}\raggedright
n
\end{minipage} & \begin{minipage}[b]{\linewidth}\raggedright
entropy
\end{minipage} & \begin{minipage}[b]{\linewidth}\raggedright
no start on the copy
\end{minipage} & \begin{minipage}[b]{\linewidth}\raggedright
L8
\end{minipage} & \begin{minipage}[b]{\linewidth}\raggedright
SIV+PO
\end{minipage} & \begin{minipage}[b]{\linewidth}\raggedright
SIV+PO\textless620
\end{minipage} \\
\midrule\noalign{}
\endhead
\bottomrule\noalign{}
\endlastfoot
O6b, 4-digit ticket, default (27.8\%) & 600 & 13.5 & 88.3 & 87.2 & 89.2
& -- \\
O6b, 4-digit ticket, 10\% (11.0\%) & 600 & 55.2 & 69.7 & 69.0 & 69.5 &
-- \\
O6b, 6-digit ID, default (27.4\%) & 100 & 43.0 & 97.0 & 96.0 & 96.0 &
-- \\
O6b, mixed ID (\texttt{QX7-4821}), default (27.7\%) & 100 & 66.0 & 94.0
& 95.0 & 94.0 & -- \\
O6b, 4-digit ticket, as released, default (30.4\%) & 600 & 19.5 & 69.3 &
76.3 & 69.7 & -- \\
O6c, 4-digit ticket, as released, new texts, default (30.2\%) & 600 &
24.2 & 66.5 & 74.2 & 66.7 & 77.3 \\
\end{longtable}
}
}

\textbf{O6b: the hazard and a gated veto.} Removing entropy's starts on
the copy raises exact copies by +74.8 points at the default rate (+71.3
to +78.2) and by +14.5 at 10\% (+11.7 to +17.5); L8 gains +73.7 (+70.2
to +77.2) and +13.8 (+10.8 to +17.0). Its costs against entropy, at the
default rate and at 10\%: computed values +0.2 (-0.5 to +0.9) and +0.4
(+0.1 to +0.8) points; final answers -0.8 (-1.7 to -0.2) and +0.3 (-0.8
to +1.5); copied numbers 0.0 (-0.4 to +0.4) and 0.0 (-0.3 to +0.4); bits
per byte outside the copy +0.0006 (+0.0002 to +0.0010) and -0.0026
(-0.0034 to -0.0018). On the same 600 problems without the ticket, L8
removed none of the 1,583 and 249 starts that entropy places at computed
values. A positive control (a veto of every start after a recurring
8-byte context) shows a cost under the same bars (computed values -5.2
points, -6.8 to -3.7). Letter IDs and a copy line reworded mid-line
showed no hazard and were not tested. The filed readings, quoted whole
(F1 is the 4-digit ticket cell, D the default rate, c1 the copy's first
byte, F5 the cell under the conversion as released; ``about 1,700'' is
the rule's expected count of entropy's computed-value starts, which were
1,583):

\begin{enumerate}
\def\labelenumi{\arabic{enumi}.}
\tightlist
\item
  ``On fresh GSM8K problems in O6's
  \texttt{\textbackslash{}nTicket\ NNNN} template, in BLT-1B with its
  512-byte local windows, scored teacher-forced, L8 restores verbatim
  copies at BLT-1B's default patch rate and at entropy@10\%, each
  compared at a matched rate. At the default rate it gains +73.7
  {[}+70.2, +77.2{]} points of exact copies, 0.98 of what removing every
  copy start gains; at 10\% it gains +13.8 {[}+10.8, +17.0{]}. Its costs
  stay within the bars, read on point estimates: at most 2 points on
  computed values, final answers and copied numbers; at most +0.01 bits
  per byte outside the copy; a patch rate within 0.005; at most 4
  new-value removals of about 1,700 computed-value starts at the default
  rate, and none at 10\%.''
\item
  ``In BLT-1B with its 512-byte local windows, the fix also holds for
  6-digit IDs and mixed IDs in the same \texttt{\textbackslash{}nTicket}
  template, at the default rate.''
\item
  Paired verdict inside the windows, ``SIV+PO better'' (L8 minus SIV+PO
  -2.0 points in F1 at D, -3.3 to -0.8; the other class rows tie): ``L8
  lost copies the byte rule kept, in F1 at D: in 5 of the 13 lost copies
  L8 did not fire at c1 (copies L8 missed); in 8 it fired, and the loss
  is a side effect of the refit.''
\item
  Paired verdict under the conversion as released, ``L8 better'' (L8
  minus SIV+PO +6.7, +4.0 to +9.5): ``Relative to SIV+PO, a rule that
  stops where the model cannot copy helps under the released conversion;
  F5's exact copies by text length show where. A byte rule with a length
  cutoff might match it; that is untested.''
\item
  ``No outcome of this test shows that reading the model is needed.''
\end{enumerate}

Under the released conversion the difference comes from long texts: in
the 79 texts of 700 tokens or more, removing entropy's starts on the
copy drops exact copies from 46.8\% to 0.0\%; L8, which fires at the
copy's first byte in 1.3\% of them, keeps 45.6\%, and SIV+PO 0.0\%.
Under 600 tokens (448 texts) L8 and SIV+PO both keep 87.9\%.

\textbf{O6c: a byte rule with a cutoff.} O6c scored the rival that
reading 4 names, on 600 test texts new by GSM8K index (problems 6,557 to
7,176; fit texts 6,253 to 6,556), under the conversion as released, at
BLT-1B's default rate only. Its class is ``the cutoff better'', with
this reading, quoted whole: ``A byte rule that stops at token 620, a
position fixed from dev texts, beats L8 under the released conversion,
on this template.'' L8 minus SIV+PO\textless620 is -3.17 points of exact
copies (-5.00 to -1.33; 25 copies lost and 6 kept; L8 did not fire at
the copy's first byte on 22 of the 25 and fired on 3). Beside it, L8
minus SIV+PO is +7.50 (+4.33 to +10.67), ``L8 better'' under O6b's rule,
as in O6b's released row; removing entropy's copy starts gains +42.33
(+37.00 to +47.67), and SIV+PO\textless620 gains +53.17 over entropy
(+49.17 to +57.17). The rule had predicted a tie. The independent check
found the run and its reading as filed and attached these
qualifications, which travel with the result:

\begin{itemize}
\tightlist
\item
  \emph{Exact copies only.} On bits per copy, the other copy readout of
  the same output, L8 and the cutoff show no detectable difference
  (-0.014 bits, -0.074 to +0.044; read after the data).
\item
  \emph{The default rate only} (entropy's test rate 30.2\%); nothing was
  run at any other rate.
\item
  \emph{One placement.} The cutoff, 620, was set by the author from the
  development texts after predictions based on O6b's released-conversion
  results had been seen, and it sits on these texts' own optimum (read
  after the data, cutoffs from 617 to 621 do best). Only this cutoff,
  and only SIV+PO, were tested.
\item
  \emph{Strength.} About 9\% of bootstrap replications of the whole test
  would tie.
\item
  \emph{New by index.} One test text asks the question of an O6b text
  with a renamed person; dropping it leaves -3.17.
\item
  \emph{Costs.} O6c's rule has no rows for costs, so it licenses no
  claim about either rule's costs.
\end{itemize}

\textbf{What the copy tests cannot show.} Free generation (every readout
is teacher-forced); real identifiers (the IDs are synthetic); other
models (only BLT-1B's base weights through this conversion); other
templates or natural text; why a start inside a copy breaks it; or
whether a copy veto would help a model trained with it. By bytes alone,
on Table 4's 293 development problems (after the data), the novelty
table starts at no copy onset at either rate only because the same two
bytes precede every ID in this template, and it scores a copy exactly as
its source, so this is not copy detection; it still starts inside 20.8\%
and 9.2\% of copies at the default rate and at 10\%.

\section{Appendix C. Outside computed values in
math}\label{appendix-c.-outside-computed-values-in-math}

\textbf{Training from scratch} (12M, 2 seeds per rule, all on Kaggle T4
GPUs in fp16; version 1's runs). We repeated the comparison on prose
(WikiText-103), Python source code (0.59 GB, held out by file) and
synthetic execution traces (0.31 GB). Both rules are fitted on the
domain's training text, and the table's threshold is set to match
entropy's evaluation patch rate (Table C1). With two seeds per rule no
difference is distinguishable, and the Welch intervals are wide (t =
12.71 or 4.30). On prose the table's bits per byte are lower on average
(1.770 against 1.832; -0.063, -0.547 to +0.422). On code, at matched
evaluation rates (7.6\% and 7.7\%), entropy's are lower on average
(1.649 against 1.684; +0.035, -0.030 to +0.100); in the higher-budget
pair the table evaluates at 13.1\% of bytes against entropy's 11.7\%, so
that pair is not matched. Whole-token exact match on code, from
separately trained runs, shows no detectable difference (identifiers not
seen in the previous 128 bytes +0.9 points, -0.8 to +2.5; copied values
after \texttt{=}, \texttt{return} or \texttt{(} -1.8, -4.1 to +0.5). On
traces the rules are not distinguishable on computed values (-4.9, -82.5
to +72.6) or copied values (+4.1, -53.7 to +62.0). In traces the same
\texttt{-\textgreater{}} precedes computed values
(\texttt{d\ =\ c\ *\ 2\ -\textgreater{}\ 44}) and copied ones
(\texttt{c\ =\ b\ -\textgreater{}\ 22}), and the two-byte table starts a
patch at none of either and at every line start
(\texttt{results/generic/trace/novnet.json}, after the data): it learns
cues that precede new content on average in its fitting text, not every
fixed cue. A 128-byte network trained to predict the same novelty target
places patch starts at 74\% of computed trace values and none of the
copied ones, yet the model trained with it scores 47.9\% on computed
values and 36.6\% on copied ones, against 50.2\% and 61.0\% with the
table.

\needspace{15\baselineskip}

\textbf{Table C1.} Outside math (12M, 2 seeds, T4 in fp16; mean ± sd).
Exact match (\%) on traces only. Differences with seed-level intervals
are in the text (\texttt{paper2/v2/results/v2\_outside\_math.json}).

{\def\LTcaptype{none} 
{\footnotesize\setlength{\tabcolsep}{3pt}
\begin{longtable}[]{@{}
  >{\raggedright\arraybackslash}p{(\linewidth - 10\tabcolsep) * \real{0.0822}}
  >{\raggedright\arraybackslash}p{(\linewidth - 10\tabcolsep) * \real{0.3973}}
  >{\raggedright\arraybackslash}p{(\linewidth - 10\tabcolsep) * \real{0.0685}}
  >{\raggedright\arraybackslash}p{(\linewidth - 10\tabcolsep) * \real{0.1781}}
  >{\raggedright\arraybackslash}p{(\linewidth - 10\tabcolsep) * \real{0.1370}}
  >{\raggedright\arraybackslash}p{(\linewidth - 10\tabcolsep) * \real{0.1370}}@{}}
\toprule\noalign{}
\begin{minipage}[b]{\linewidth}\raggedright
domain
\end{minipage} & \begin{minipage}[b]{\linewidth}\raggedright
rule
\end{minipage} & \begin{minipage}[b]{\linewidth}\raggedright
eval patch rate
\end{minipage} & \begin{minipage}[b]{\linewidth}\raggedright
bits/byte
\end{minipage} & \begin{minipage}[b]{\linewidth}\raggedright
computed values
\end{minipage} & \begin{minipage}[b]{\linewidth}\raggedright
copied values
\end{minipage} \\
\midrule\noalign{}
\endhead
\bottomrule\noalign{}
\endlastfoot
prose & entropy & 10.1\% & 1.832 ± 0.031 & -- & -- \\
prose & novelty table & 10.3\% & 1.770 ± 0.044 & -- & -- \\
code & entropy & 7.6\% & 1.649 ± 0.004 & -- & -- \\
code & novelty table & 7.7\% & 1.684 ± 0.006 & -- & -- \\
code & entropy (higher budget) & 11.7\% & 1.566 ± 0.021 & -- & -- \\
code & novelty table (higher budget) & 13.1\% & 1.541 ± 0.005 & -- &
-- \\
traces & entropy & 10.0\% & 1.127 ± 0.033 & 55.1 ± 8.6 & 56.9 ± 5.4 \\
traces & novelty table & 10.3\% & 1.157 ± 0.014 & 50.2 ± 0.2 & 61.0 ±
3.5 \\
traces & novelty network (128 bytes) & 10.0\% & 1.196 ± 0.001 & 47.9 ±
0.3 & 36.6 ± 0.3 \\
\end{longtable}
}
}

\textbf{Where an extra start pays, by domain, at inference} (rules fixed
beforehand: \texttt{rules/XDOMAIN\_RULES.md},
\texttt{R11\_PROBE\_XDOMAIN\_RULES.md},
\texttt{R15\_NEWTOKEN\_XDOMAIN\_RULES.md}). On BLT-1B, in six domains
with 60 test documents each (GSM8K solutions, MATH algebra, Python
standard-library source, English Wikipedia, MedQA clinical vignettes,
synthetic traces), each rule adds an extra 3\% of patch starts to
entropy's layout at 10\%. We measured the gain of a start at every
position on two objectives: bits over the next 16 bytes, and exact match
of the whole word, number or identifier that starts there (a unit).
Gains were measured with 1/48 of positions toggled at once; against
single starts (12 positions in one document each) the grouped gain
correlates 0.88 in math and 0.93 in prose. A rule's score is the share
of an oracle's gain it captures; differences are in that share, with
intervals over documents (1,000 resamples).

\begin{itemize}
\tightlist
\item
  Entropy plus the table beats entropy on units only in code (+0.08,
  +0.01 to +0.17) and is below it in prose (-0.09, -0.16 to -0.03). In
  math it is not distinguishable (+0.02, -0.04 to +0.09). A possible
  reason, not tested, is that computed values are a small share of all
  units in GSM8K text and this test adds 3\% on top of a 10\% layout, so
  the large gain on computed values of §6 need not show in the share
  over all units.
\item
  The table alone is below entropy on bits in four domains (algebra
  -0.12, code -0.06, prose -0.12, clinical -0.17; each interval below
  0).
\item
  A two-byte map of measured gains, fitted on the other five domains,
  beats entropy in no domain on either objective; fitted in-domain it
  does on bits in math (+0.10), algebra (+0.05) and traces (+0.05; each
  interval above 0). A map fitted on math does not carry to algebra on
  units (-0.07, -0.15 to +0.01), nor does the table (-0.07, -0.17 to
  +0.03).
\item
  A probe on BLT-1B's patcher state trained on measured gains beats
  entropy on bits within five domains (+0.03 to +0.13, each interval
  above 0) but on units only in traces; fitted on the other domains, it
  beats entropy only on bits in traces.
\item
  A label-free probe for ``a new word or number starts here'' detects
  its own label well (AUC 0.93 to 0.98) but beats entropy on units in no
  domain and on bits only in code and traces; in math it is below the
  probe for new numbers alone (-0.11 on bits, -0.14 to -0.07).
\end{itemize}

In math, values after \texttt{=} carry the highest unit gain of any byte
class (0.42 per start), and only 14\% of them are in entropy's layout:
this is the case the paper studies. No rule tested beat entropy on units
in more than one domain, on one model, at inference only. Exploratory,
with no verdict set: a structure-only rule that adds starts at word and
number starts, ranked by entropy, beats entropy's next picks on bits in
five of six domains (+0.02 to +0.10, each interval above 0; traces
+0.012, -0.003 to +0.027). This is consistent with SpaceByte's
word-boundary rule \citep{slagle2024spacebyte}, which is prior art; on
units it equals entropy by construction, since the unit objective
already restricts every rule to unit starts and this rule ranks them by
entropy.

\section{Appendix D. Other negative
results}\label{appendix-d.-other-negative-results}

\textbf{Classic compressors as patch rules.} If LZ novelty were the
right target, a stronger compressor might be a better rule. It is not:
on BLT-1B at 10\%, document-level LZ (a patch start where the text so
far has no recent match) gets 15.1\% of computed results and PPM-C's
escape probability 13.4\%, both below entropy (-3.9 points, -7.0 to
-1.0; -5.5, -8.7 to -2.6); they start a patch at only 7\% and 10\% of
computed results. Adding LZ to the table does not improve it (54.1\%
against 56.5\%; -2.4, -4.9 to +0.1).

\textbf{Value uncertainty.} A rule that starts a patch where the patcher
is sure of the kind of the next byte (digit, letter, punctuation) but
unsure of its value finds computed results at inference on BLT-1B
(48.7\% at 10\%, +29.8 points over entropy, +25.4 to +33.9) but loses
final answers there (32.0\% against 51.4\%). As a training rule it fails
at 12M (final answers 18.9\% against 46.6\% for entropy; -27.7, -43.2 to
-12.2; value uncertainty on T4s in fp16, entropy on the A40 in bf16). At
50M, on one device, it is below entropy on average (35.1\% against
59.9\%; -24.9, -77.3 to +27.5), with one seed at 11.1\%, so the
difference is not distinguishable. At 1M it is near the floor, as is
entropy (final answers 3.9\% against 7.0\%; -3.1, -8.3 to +2.1; across
devices). With a forced start after the answer marker, its final answers
at 50M rise to 65.7\% while computed results stay near 9\% (9.5\%,
against 8.9\% without the forced start).

\textbf{Switching rules after training.} In two separate T4 re-runs of
12M table seeds 0 and 1 (final answers 68.9\% and 78.3\% under their own
layout), changing the rule at inference lowers final answers to 0.0\%
under entropy and 0.5\% under value uncertainty (means over the two
runs), and raises bits per byte from 1.66 to 2.5-2.9. A model trained
under a layout learns to use it.

\section{Appendix E. A one-seed training pilot of the probe
hybrid}\label{appendix-e.-a-one-seed-training-pilot-of-the-probe-hybrid}

One seed, exploratory. A 12M model was trained on an Apple laptop GPU
(MPS, fp32) under entropy plus the top 3\% of a label-free probe on the
training-time patcher (trained on auto-labels, as in Appendix A), with
the corpus, recipe and evaluation of the 12M runs of Table 1. It got
73.5\% of final answers, 10.5\% of computed results and 1.614 bits per
byte, which the rule written before it ran reads as ``promising'' (at
least 60\% of final answers with at most 1.70 bits per byte). On final
answers that is above every 12M entropy seed (42.0-49.1\%) and table
seed (70.9-72.3\%) and just below the lowest seed of the hand-written
rule (73.9\%); its bits per byte are below every run of those arms
(1.622-1.707). It is not a test: one seed, on another device and
precision, at an evaluation patch rate of 13.0\% against entropy's
11.2\%, and on text with the \texttt{=} and
\texttt{\textgreater{}\textgreater{}} cues, where every cue rule beats
entropy on final answers at 12M (the table only in a comparison across
devices). The planned test (three seeds with a decision rule;
\texttt{rules/R13\_PROBE\_TRAIN\_RULES.md}) was not run, and by its own
amendment even a pass would have shown only that the probe works as a
training rule on text with cues.

\section{Appendix F. Work that was not
run}\label{appendix-f.-work-that-was-not-run}

Designed with decision rules and not run when the line of work closed on
2026-11-01 (no results exist):

\begin{itemize}
\tightlist
\item
  \textbf{Study A beyond its interim look} (145M, the table against
  entropy, 3 seeds; \texttt{prereg/2026-10-13\_scale150\_bltft.md},
  \texttt{paper2/v2/rules/STUDY\_A\_INTERIM\_RULES.md}): seeds 1 and 2
  never finished, and no prediction was evaluated (below).
\item
  \textbf{Study B} (BLT-1B fully fine-tuned at a 10\% budget, the table
  against entropy; same preregistration).
\item
  \textbf{The hand-written rule at 50M} with the table's recipe
  (clipped, 3 seeds, A40; final answers decisive;
  \texttt{rules/R2\_DECISION\_RULE.md}).
\item
  \textbf{A 12M training budget curve} (entropy at 15\% and 20\%, the
  table at 15\%) with a 12M table run on the A40 as a same-device anchor
  (\texttt{rules/R8b\_HYBRID\_D\_RULES.md}, part D). The size of the
  device effect at 12M is therefore unknown.
\item
  \textbf{The label-free hybrid of §6 trained from scratch} (12M, 3
  seeds; \texttt{rules/R8c\_HYBRID\_TRAIN\_RULES.md}).
\item
  \textbf{The probe hybrid trained from scratch} (12M, 3 seeds;
  \texttt{rules/R13\_PROBE\_TRAIN\_RULES.md}); only the pilot of
  Appendix E ran.
\end{itemize}

Planned, with no decision rules written: the detector on prose from a
second writer, and a confirmation of §5's cue split on newly written
prose (every prose rewrite here comes from one writer); copy checks on
BLT-7B and on H-Net at a controlled budget.

Never run and untested: a table fitted on OpenWebMath alone;
word-boundary (SpaceByte) or fixed-stride rules trained from scratch; a
hand-written rule with boundaries at line starts (Appendix H); entropy
plus boundary dependence on bits per byte; the detector on other problem
sources, writers or models; other cutoffs, byte rules or rates in the
copy tests; free generation for any result.

\textbf{Study A's interim look.} Study A was preregistered on 2026-10-13
(\texttt{prereg/2026-10-13\_scale150\_bltft.md}, where the size is
called ``150M''; the model has 145M parameters) to compare the table
with entropy, boundary dependence and the corpus-fitted table at 145M
parameters, 3 seeds each, 91,500 updates. Amendment A-4, which is not in
the release and is summarized in
\texttt{paper2/v2/rules/STUDY\_A\_INTERIM\_RULES.md}, reduced it to the
table against entropy, 3 seeds each, on Kaggle T4 GPUs in fp16 (against
the A40 in bf16 at 50M). Its predictions were then three: final answers
at least 10 points higher; a 145M final-answer gap no more than 5 points
below the 50M gap; and bits per byte within +0.03. The predictions that
needed the other arms (against dependence, and on the corpus-fitted
table) were not run. A gate at 50M that had to pass before money was
spent (each table seed at least 10 points above dependence on final
answers and on computed results) was missed by 0.3 points on one seed,
and the study went ahead, as its record discloses. It was stopped on
2026-11-01, when this line of work closed, before seeds 1 and 2 had
finished, so it has no verdicts. Its only reported result is an interim
look at seed 0 (amendment A-5), added before any Study A result was
opened and allowed only to stop the study. For the table against
entropy: final answers 91.2\% against 68.9\% (+22.3 points), computed
results 68.3\% against 29.3\% (+39.0), bits per byte 1.813 against
1.492, all teacher-forced. Both arms' bits per byte were worse than
every 50M run, which fired the look's rule for broken training. A later
diagnosis on the part-trained second seed (a disclosed deviation: the
rule had kept it closed) found memorization of the GSM8K training
problems that the corpus repeats (0.84 bits per byte on training
problems against 1.57 on new ones), which would account for seed 0's
worse bits per byte, and a divergence in fp16 of one entropy run, rolled
back by amendment A-6. These values are recorded in the rules file; the
diagnostic output is not released. The interim rule said seed 0 would
not be reported before all three seeds were in; we report it because the
other seeds will not exist. It is one seed, interim and not
confirmatory, on other hardware than 50M, with bits per byte confounded
by memorization: it neither supports nor rules out growth with size, and
Study A has no hand-written arm.

\section{Appendix G. A check of public H-Net
checkpoints}\label{appendix-g.-a-check-of-public-h-net-checkpoints}

Version 1 stated that H-Net's outer chunks start at the \texttt{=}
before 100\% of in-line computed results and at 100\% of final answers,
and attributed this to \citet{vera2026easy}, which does not contain it.
The measurement comes from a check of the public H-Net checkpoints
\citep{hwang2025hnet} not reported before
(\texttt{scale/cloud/hnet\_check.py}; data in \texttt{results/hnet/}),
at their native chunk rates with their boundaries as trained, on 1,292
GSM8K test solutions without annotations (3,547 in-line computed
results). In the 2-stage model, the outer stage (a chunk start at 18.4\%
of bytes) starts a chunk at the \texttt{=} two bytes before 99.9\% of
the computed results, and at the result's own first byte for 0.06\% of
them; it starts a chunk at the space before 99.8\% of final answers. In
the 1-stage model (27.0\% of bytes), a chunk starts at a result's first
byte for 23.9\% of computed results (all of them, counting the preceding
space), and the results with such a start were less accurate than the
others (teacher-forced exact match 36.6\% of 849 against 51.2\% of
2,698; correlational). At these native rates, well above the 10-15\%
budgets studied here, the check says nothing about tight budgets, and it
does not show that learned chunkers find such structure in general.
Version 1's H-Net numbers on Python source are withdrawn with it.

\section{Appendix H. Where each rule spends its budget (after the
data)}\label{appendix-h.-where-each-rule-spends-its-budget-after-the-data}

\textbf{Coverage, not enrichment} (Table H1). Boundary enrichment
\citep{neitemeier2026sombrero} is the mean surprisal at patch starts
over the mean over all bytes, with surprisal from the small model that
drives entropy patching. On the math evaluation set at the 50M
thresholds, entropy has the highest enrichment (\(B = 1.61\)) and starts
a patch at 0.7\% of computed-result starts; the table sits close to
chance (\(B = 1.13\)) and covers all of them. The hand-written rule has
high enrichment (\(B = 1.52\)) and also covers all of them, and at 12M
it beats entropy on both targets (§4): what goes with accuracy is
coverage of value starts, not low enrichment. The small patcher's mean
surprisal is 3.34 bits at computed-result starts and 5.39 at
final-answer starts.

\needspace{11\baselineskip}

\textbf{Table H1.} Boundary enrichment and coverage on the math
evaluation set at the 50M thresholds, with 50M teacher-forced exact
match (\%; clipped, 3 seeds, except the hand-written rule: unclipped, 2
seeds).

{\def\LTcaptype{none} 
{\footnotesize\setlength{\tabcolsep}{3pt}
\begin{longtable}[]{@{}
  >{\raggedright\arraybackslash}p{(\linewidth - 12\tabcolsep) * \real{0.3065}}
  >{\raggedright\arraybackslash}p{(\linewidth - 12\tabcolsep) * \real{0.0806}}
  >{\raggedright\arraybackslash}p{(\linewidth - 12\tabcolsep) * \real{0.1613}}
  >{\raggedright\arraybackslash}p{(\linewidth - 12\tabcolsep) * \real{0.1290}}
  >{\raggedright\arraybackslash}p{(\linewidth - 12\tabcolsep) * \real{0.1129}}
  >{\raggedright\arraybackslash}p{(\linewidth - 12\tabcolsep) * \real{0.1290}}
  >{\raggedright\arraybackslash}p{(\linewidth - 12\tabcolsep) * \real{0.0806}}@{}}
\toprule\noalign{}
\begin{minipage}[b]{\linewidth}\raggedright
rule
\end{minipage} & \begin{minipage}[b]{\linewidth}\raggedright
eval rate
\end{minipage} & \begin{minipage}[b]{\linewidth}\raggedright
enrichment B
\end{minipage} & \begin{minipage}[b]{\linewidth}\raggedright
computed starts covered
\end{minipage} & \begin{minipage}[b]{\linewidth}\raggedright
final starts covered
\end{minipage} & \begin{minipage}[b]{\linewidth}\raggedright
computed
\end{minipage} & \begin{minipage}[b]{\linewidth}\raggedright
final
\end{minipage} \\
\midrule\noalign{}
\endhead
\bottomrule\noalign{}
\endlastfoot
entropy & 11.2\% & 1.61 & 0.7\% & 100\% & 20.7 & 59.9 \\
dependence & 10.9\% & 1.02 & 36.7\% & 0\% & 35.1 & 78.2 \\
hand-written rule & 12.1\% & 1.52 & 100\% & 100\% & 43.4 & 85.5 \\
novelty table & 11.9\% & 1.13 & 100\% & 100\% & 57.5 & 89.5 \\
corpus-fitted table & 11.4\% & 1.03 & 99.3\% & 100\% & 57.3 & 86.4 \\
\end{longtable}
}
}

\textbf{What each rule selects.} On BLT-1B (1,292 GSM8K test problems),
we placed each rule's top 10\% of positions on two axes: how surprising
the position is to BLT-1B itself, and whether the word or number there
is new in the problem. Entropy's selections sit at BLT-1B's 85.7th
surprisal percentile and the table's at the 59.6th (difference 26.0
points, 25.7 to 26.3; 400 problem resamples); computed-result starts sit
at the 63rd and word starts at the 75th. These are percentiles of
BLT-1B's own surprisal, not of the patcher's entropy used in §5. Both
rules pick new content at similar rates: 64.2\% and 65.3\% of their
selections are new words or numbers, against 49.3\% of all positions
(the table slightly more, +1.2 points, +0.6 to +1.7; +6.7 among numbers,
+5.7 to +7.7). An independent LZ measure ranks positions only weakly
like either rule (Spearman 0.10 for the table, 0.19 for entropy), and
its own top 10\% covers 1.0\% of computed-result starts; it is one minus
the share of the next 8 bytes that lie inside an LZ77 match with
anything earlier in the problem, where the table's target \(m_t\) is the
length of the match that starts at the position, within the previous 127
bytes. This is descriptive, and partly circular, because the small and
large models' surprisals are correlated; it does not identify a cause.
It does not support version 1's title, ``Patch where the text is new''.

\textbf{Segment starts (a hypothesis).} On held-out math text in the
training format, the table (at a 9.5\% rate) starts a patch at 75\% of
line starts, which are 9\% of its boundaries against 1\% of bytes; in
code it starts a patch at none of the statement starts, and 57\% of its
boundaries fall inside identifiers (50\% of bytes). On traces, the
128-byte novelty network of Appendix C placed starts at 74\% of computed
values and none of the copied ones, yet the model trained with it was
worse than with the table. These are consistent with the table helping
in training partly through boundaries at segment starts. We did not test
this: a 12M hand-written rule with boundaries at line starts would, and
it was not run.

\end{document}